# Smart Automotive Technology Adherence to the Law

## (De)Constructing Road Rules for Autonomous System Development, Verification and Safety

Scott McLachlan, Martin Neil, Kudakwashe Dube, Ronny Bogani, Norman Fenton, and Burkhard Schaffer

Risk and Information Management, Queen Mary University of London, London, UK
Birmingham Law School, University of Birmingham, Birmingham, UK
SCRIPT Centre, Edinburgh Law School, University of Edinburgh, Edinburgh, UK
School of Fundamental Sciences, Massey University, Palmerston North, NZ
Health informatics and Knowledge Engineering Research (HiKER) Group

**Abstract**

Driving is an intuitive task that requires skills, constant alertness and vigilance for unexpected events. The driving task also requires long concentration spans focusing on the entire task for prolonged periods, and sophisticated negotiation skills with other road users, including wild animals. These requirements are particularly important when approaching *intersections, overtaking, giving way, merging, turning* and while adhering to the *vast body of road rules.* Modern motor vehicles now include an array of smart assistive and autonomous driving systems capable of subsuming some, most, or in limited cases, all of the driving task. The UK Department of Transport's response to the *Safe Use of Automated Lane Keeping System* consultation proposes that these systems are tested for compliance with relevant traffic rules. Building these smart automotive systems requires software developers with highly technical software engineering skills, and now a lawyer's in-depth knowledge of traffic legislation as well. These skills are required to ensure the systems are able to safely perform their tasks while being observant of the law. This paper presents an approach for deconstructing the complicated legalese of traffic law and representing its requirements and flow. The approach (de)constructs road rules in legal terminology and specifies them in *structured English logic* that is expressed as *Boolean logic* for automation and *Lawmaps* for visualisation. We demonstrate an example using these tools leading to the construction and validation of a *Bayesian Network model.* We strongly believe these tools to be approachable by programmers and the general public, and capable of use in developing Artificial Intelligence to underpin motor vehicle smart systems, and in validation to ensure these systems are considerate of the law when making decisions.

# 1. Introduction

When powered vehicles first hit the roads at the beginning of the 20th century, they brought in their wake major changes to the law. Road traffic law emerged as a legal field in its own right, with complex rules that made it difficult for citizens to conduct their business in a lawful manner, potentially turning everyone (or at least everyone driving) into a criminal, and as a result, assigning significant discretionary power to our police. Policing also changed dramatically, requiring new laws, new organisational structures, and new procedures. To give an example, the only police force in the UK that is permitted to enforce the law across all of England and Scotland, is the British Transport Police. The legal ramifications were felt in all parts of society, and by all people: whether drivers or not. Some jurisdictions introduced new offences such as jaywalking that resolved conflicts about control of public spaces: non-drivers had to give up historical rights to allow an inherently dangerous technology to be deployed safely.

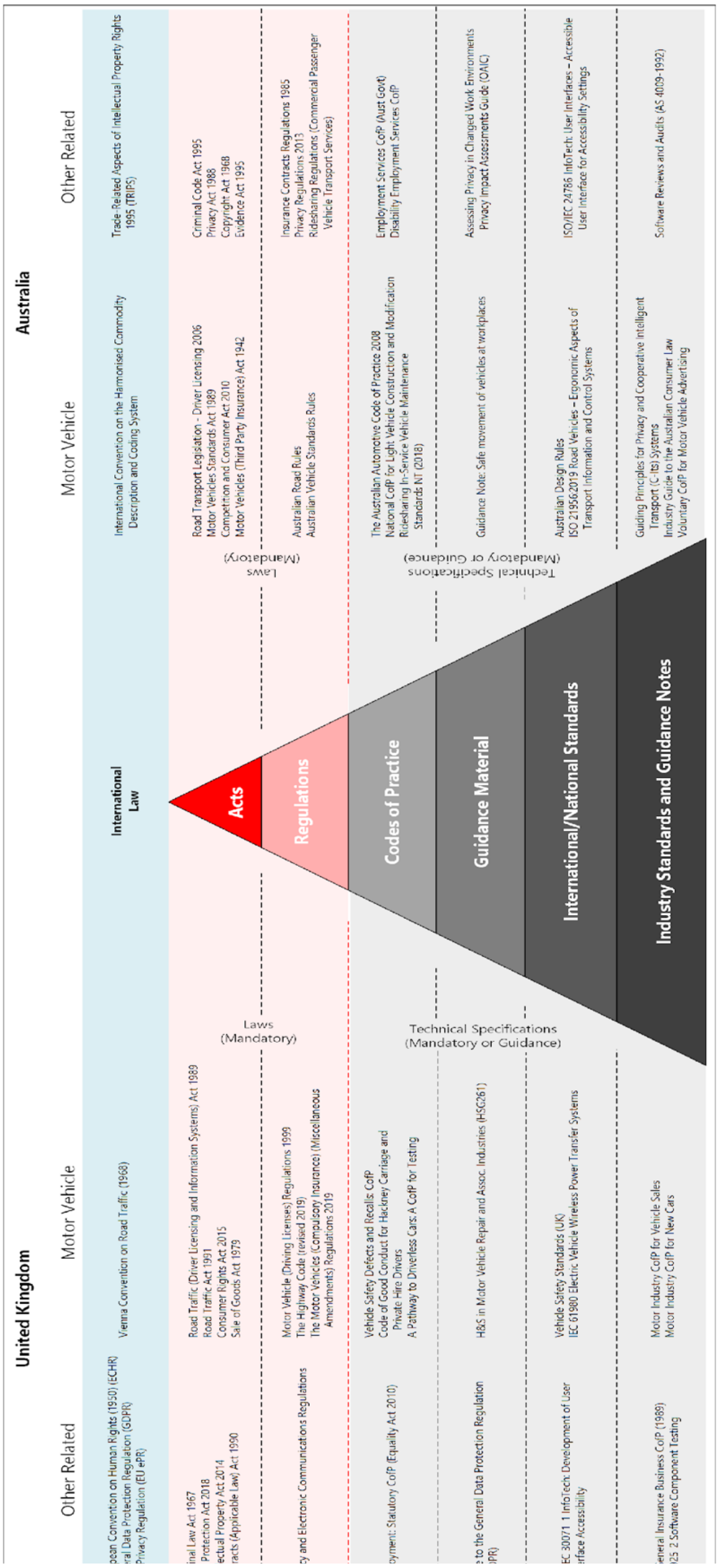

*Figure 1: Motor vehicle laws, regulations and standards*

Today a wide array of legislation, regulation and standards ascribe complex legal responsibilities on manufacturers, importers and distributors, owners, and vehicle users. Rules impact elements of the design, materials and manufacturing processes. They set minimum standards for safety and protection of occupants and other road users in actual and potential accident situations. Soon, two complementary sets of rules evolved, One set is directed at the manufacturer of cars, the other at their drivers. In Figure 1, these are represented by the bottom three categories. The other set of rules is directed at the driver and can take the form of laws such as the Road Traffic Act, but also intersect with rules not specific to the domain such as criminal and civil liability laws. This separation of responsibilities benefits both, drivers and manufacturers. Take the example of *brakes*. Their design and manufacturing is covered by a detailed set of rules including the succinctly named: *Commission Directive 85/647/EEC of 23 December 1985 adapting to technical progress Council Directive 71/320/EEC on the approximation of the laws of the Member States relating to the braking devices of certain categories of motor vehicles and their trailers*[1]. It is the shared duty of the designer, the supplier who manufactures the brakes, and the vehicle manufacturer who integrates them into the car to know and adhere to these rules. But this is also where their responsibility stops. They need not ensure that the driver actually uses the brakes when, and only when, it is appropriate. Drivers are in turn subject to road traffic law that requires they drive with *due care and attention*. This includes a duty to brake when necessary in order to prevent a collision, and not to brake unnecessarily when it may result in one (so not randomly when driving at speed on the motorway for instance). The driver's task is to make correct and appropriate decisions regarding when to brake, without a requirement to understand the minute details of how their brakes work or whether they can rely on them.

[1] https://eur-lex.europa.eu/legal-content/EN/ALL/?uri=CELEX%3A31985L0647

This picture is admittedly oversimplified. The driver or owner has duties to ensure the roadworthiness of their cars, exemplified in a requirement to submit the car for regular inspection[2]. Such requirements act as bridging rules between the two domains. Other laws also apply across both domains, including meta-rules that determine adherence to primary rules with higher-ranking provisions such as the constitution. We will discuss below how this interconnectedness of the normative legal order poses challenges for all attempts to represent legal knowledge in a computational format.

We note already here what we call the *semantic normative gap*: while the aim of technical regulations is ultimately to enable drivers to drive lawfully – as defined by the laws directed to drivers - they do not make this connection explicit. For the norm addressee, this has several benefits. It allows them to make design decisions without necessarily having to worry about "the big picture". It simplifies the rules into easily actionable instructions that do not need interpretation. It also means that translating them into formal representations that can be understood by a computer is relatively straightforward, certainly far more easily than any other field of law, and they embody the formalist ideal of law as a system of clear and unambiguous rules that are self-applying. By contrast, laws directed at drivers typically use vague terms that are considerably more flexible than the technical guidelines and standards imposed on manufacturers, such as *reasonable care and attention*, and are therefore able to respond to unforeseen circumstances and involve balancing between often competing values. Their correct application requires considerable contextual knowledge not just about the physical but also the social world. As a consequence, they are much more difficult to capture formally.

Secondly, we note that the demarcation between the domains was contested from the very beginning. A simple example is the choice between laws against speeding directed solely at the driver, and the design requirement to limit the highest possible speed a car can achieve to the maximum permitted in a given jurisdiction[3]: an early example of *compliance through architecture*. Very early on in the history of the car we find even more sophisticated suggestions were made. For example, Charles Adler embedded magnetic plates in the road which would slow down any vehicle driving over it to 24km/h by activating the vehicle's speed governor[4].

[2] Known as *MOT Tests* in the United Kingdom, *Warrant of Fitness* or WOF checks in New Zealand, or a *Roadworthiness Check* in Australia.
[3] For example, the legal definition of a *moped* in the United Kingdom was revised in 1977 to include a maximum design speed of 30mph (revised up in the 1990's to 50mph and later again in the late 2000's back down to 45mph) and requiring manufacturers to fit a speed restriction device, or governor, to ensure the machine as manufactured could not exceed the prescribed speed and therefore fall outside of the legal definition.

As more and more parts of the car became "smart" and driven by software, the paradigm of "compliance by architecture" or "software code as law" becomes more and more feasible, and reaches its pinnacle in semi- or fully automated cars. The idea of software code as a form of regulation became popular in the wake of Lessig's "pathetic dot" and the idea of "law as code" in the late 1990s. While Lessig does not require the software code be isomorphic to legal rules to be considered "coded law", the idea was quickly picked up by the law and AI community that developed explicit formal representations of applicable law to ensure for instance that electronic agents behave in accordance with contract law. Wendell Wallach and Colin Allen "Moral Machines: Teaching Robots Right from Wrong" from 2009 was another milestone. Compliance software did indeed become one of the commercial success stories of legal AI.

Some more recent papers have already applied this idea to autonomous cars, most notably Henry Prakken's formalisation of the Dutch Traffic Code[5], or Giovanni Sator's proposal of an 'ethics knob' for cars that would allow them to resolve ethical decisions in alignment with their user's preferences[6].

While of great academic interest, these attempts to deploy legal reasoning on the side of the car while it navigates through traffic faces significant obstacles. Some of them are of a purely technical nature – the type of languages best suited to formally express legal rules are not the ones used to power smart cars, and the computational costs to reach split second decisions would be prohibitively high. Other more fundamental issues relate to the characteristics of legal language and the semantic-normative gap we identified above: Those rules most amenable to formalisation are insufficient to ensure "top level compliance" with rules directed to human drivers, while laws directed at human drivers use features tailored at human readers, including the use of vague terms, that make them difficult to formalise and also in many cases irrelevant for an automated driver.

By contrast, one of the most ambitious and interesting projects in the attempt to use legal AI to build law-compliant IoT devices is the *SmarterPrivacy* project at the KIT[7]. In their approach to build a *law compliant* smart grid energy infrastructure, the legal AI does not reside on the side of the IoT (in our case, not in the smart car) but in the design environment of the engineers that build them.

---

[4] L. Vinsel, (2016). "The smart road not taken [Tech History]," in *IEEE Spectrum*, vol. 53, no. 8, pp. 46-51, DOI: 10.1109/MSPEC.2016.7524173.
[5] Prakken, H. (2017). On the problem of making autonomous vehicles conform to traffic law. *Artif Intell Law* **25,** 341–363. https://doi.org/10.1007/s10506-017-9210-0
[6] Contissa, G., Lagioia, F., & Sartor, G. (2017). The Ethical Knob: ethically-customisable automated vehicles and the law. *Artificial Intelligence and Law*, *25*(3), 365-378.
[7] https://kawa.nazemi.net/en/projects/smarter-privacy/

We can think of them as a more intelligent form of Clippy[8] that assists the designer with legal advice. Imagine a grid developer  decides where in the smart energy grid to put an AI that looks for fraudulent patterns, and for this needs data from 3 sources. At that point the legal AI that observes the design proposals that are made pops up and says: *It looks like you are routing personal identifiable data about credit worthiness to the electricity supplier. Under Art 6.5 GDPR, this requires consent, and currently the consent forms given to the electricity supplier do not contain information about credit rating agencies, so your current solution is not compliant*. To be able to perform this function, the legal design AI has an explicit formal representation of the applicable rules of the *General Data Protection Regulation* (GDPR), together with an ontology that enables it to subsume design decisions that talk about specific objects such as credit cards or electricity under the legal rules

Our paper follows this general philosophy. It differs however in some crucial aspects from SmarterPrivacy. In SmarterPrivacy, all the design decisions are taken centrally, and the designer has full access to, and knowledge of, the model of the grid that is constructed. All design decisions furthermore are in binary form: data either *is or is not* personally identifiable, and it either *is or is not* accessible. By contrast, designers and manufacturers in the highly complex car supply chain often work under uncertainty. There are at least three possible sources for uncertainty in this context, and each will have to be discussed in this paper.

The first type of uncertainty can be caused by the very nature of AI as a probabilistic tool: the smart sensors of the car will learn and improve over time, and in ways that are not fully transparent even to the vehicle's developers. Second, the uncertainty can be caused by the interdependence of design choices across the chain. Design decisions taken locally for a specific module or sensor a supplier is responsible for can be, based on that supplier's knowledge of the use case for the component, perfectly safe and reasonable. Yet, that module can create danger when it interacts with another *equally safe and reasonable* component. We accommodate these two types of uncertainty through the choice of formalism used in this paper, which differs from the one proposed by SmarterPrivacy. On the one hand, while we start like them with a simple formalisation of laws as binary rules, we show how our system can be translated into a Bayesian network that allows reasoning with uncertainty. Second, we combine the formal representation of laws as propositions with a visualisation technique, Lawmaps. Lawmaps are a highly intuitive tool that is particularly suited for collaborative tasks. In this way, we hope to assist engineers making design decisions for their component of the autonomous vehicle. Aiding them to make explicit how their efforts may contribute to the ultimate aim of a *law complaint car*, while creating sharable digital objects that then can be augmented by groups working on adjacent tasks. Additionally, we consider these Lawmaps also as a first answer to the semantic-normative gap we mentioned above: they enable designers to see how compliance with low level standards and regulations ultimately enabled the compliance with the top level laws currently directed at human drivers. Third, uncertainty can come from the law itself. As noted above, laws directed at human drivers contain significant vagueness and open texture to accommodate unpredictable future scenarios. It is not straightforward to anticipate how the law will have to change once the driver as crystallisation point of obligations is replaced by an AI. The aim is, as the UK government expressed, to have a car that "abides by the relevant law", but due to the semantic-normative gap, it is not straightforward how compliance with laws from the set currently directed at human drivers can be subsumed by the set that have prescribed the design decisions. For some laws, the answer is simple: human drivers are currently prohibited from speeding. For self-driving cars, this means their AIs must also be able to ensure the car remains within posted speed limits. But how should we think about rules like Highway Code Rule 161 that prescribes that *all mirrors should be used effectively throughout your journey*? Mirrors are a design choice based on human requirements, and the fact that our two eyes have only a limited field of vision. For the autonomous car this rule may not be necessary as their sensors can look directly behind the vehicle or into those areas that often constitute a blind spot for us. Is it therefore still necessary to attach mirrors to the autonomous car? Finally, we will see rules where the legislator will have a choice. We will look in particular at the rule that currently obligates the driver to ensure that the seat belts of any children in the car are fastened and remain so. With the driver removed, who, if anyone, *inherits* this obligation? One answer could be that a law compliant self-driving car should check the status of the seatbelts and stop driving if these are not fastened. But we can also imagine that the duty might be transferred to the next responsible adult, even if they too are only a passenger. We will see how our approach can help to identify such choice points for the legislators, and assist the designer to plan for this eventuality even though future legal development can continue to create temporary uncertainty.

[8] Clippy was the default paperclip-shaped animated character in English language versions of the *Microsoft Office Assistant* that came bundled with versions of the Microsoft Office productivity application set from 1997-2003.

Graphical representations for *belief networks* have been applied to evidence analysis for more than a century[9], and a variety of approaches[10] have sought to map sentencing processes[11], contract timelines[12] and support juridical and lay-juror decision making at the conclusion of legal proceedings[13]. However, approaches for visually modelling or representing the processes described within core legislation tend to be rare, and those that are identified are often limited in description, repeatability or applicability[14]. Lawrence Lessig coined the term *code is law* in reference to software, or code, regulating user behaviour in cyberspace much like legislation was traditionally intended to do in the physical world. He clarified this by explaining[15] that code determines how easy it is to: (i) protect privacy; (ii) monitor user actions; (iii) censor speech; and (iv) access information, which it could be said now includes mainstream alternative voices and perspectives. All this because private actors like the leaders of social media companies can embed their morals, values and policies of the government of the day into the operating code of their platforms, effectively constraining where[16] and what we are allowed to read[17], say[18] or do[19]. The code need not be adherent to or reflective of real-world law. Simply put, in Lessig's view code already operated as a law unto itself. Until recently[20], little attention focused on imbuing artificial intelligences and autonomous systems with real-world laws, regulations and policies that could guide their decision-making processes to ensure they are aware of the requirements of the law, and approaches for verification of the application and action of those laws in the resulting autonomous system's decisions.

[9] *Wigmore, J. H. (1913). "The problem of proof". Illinois Law Review.* ***8*** *(2): 77–103*; *and* Fenton, N., Neil, M., & Lagnado, D. A. (2013*). A general structure for legal arguments about evidence using Bayesian networks.* Cognitive science, 37(1), 61-102.

[10] McLachlan, S. & Webley, L. (2021). Visualisation of law and legal process: An opportunity missed. *Information Visualisation*, 20(2-3), pp 192-204.

[11] Dhami M. A. (2013) 'decision science' perspective on the old and new sentencing guidelines in England and Wales. In: Andrew Ashworth and Julian Roberts (Eds.) Sentencing guidelines: exploring the English model. Oxford: Oxford University Press, pp.165–181.

[12] Haapio H and Passera S. (2012) Reducing contract complexity through visualization-a multi-level challenge. In: 16th international conference on information visualisation, pp.370–375. IEEE. DOI 10.1109/IV.2012.68

[13] Fang J. (2014) 12 confused men: using flowchart verdict sheets to mitigate inconsistent civil verdicts. Duke Law J: 287–331

[14] For example: V. Strahonja, "Modeling Legislation by using UML State Machine Diagrams," *2006 Canadian Conference on Electrical and Computer Engineering*, 2006, pp. 624-627, doi: 10.1109/CCECE.2006.277286; *and* Smith, P. & Schwarz, V. (1987). Logical analysis of legislation using flow diagrams. *J. of the Operational Research Soc.* 38(10), pp. 981-987.

[15] Lessig, L. (2001). Code is Law. *Harvard Magazine*, January

This work investigates one approach for deconstructing legislation and regulation textually in the form of *Boolean algebra* which can subsequently be diagrammatically represented using an information visualisation (infovis) approach known as *Lawmaps*. These tools are then demonstrated in an evaluation of the semi-autonomous and autonomous system components of three current model vehicles; investigating whether these vehicle's systems are capable of adhering to twenty-three of the United Kingdom's Road Rules.

After defining and describing the *advanced driver assistance systems* (ADAS) that are currently available, we describe the vehicles that were used in this testing. We go on to present one approach for deconstructing legislation and regulation, and describe the creation of tools that can be used to incorporate considerations of law in the development of autonomous systems, and validation of the operation of that law in the decision-making processes of those systems. Then we report on a practical experiment of the application of the (de)construction approach to developing tools for one jurisdiction's traffic rules, and the use of those tools in testing adherence to those traffic rules by the systems provided in our test vehicles. We conclude with a discussion of the results of these experiments and the impact for manufacturers, vehicle users and the law.

## 1.2. Background

The United Kingdom (UK) Department of Transport (DoT) 2015 report *The Pathway to Driverless* Cars found that *real-world testing of automated technologies is possible in the UK today, providing a test driver is present and takes responsibility for the safe operation of the vehicle; and that the vehicle can*

---

[16] Rogers, R. (2020). DePlatforming: Following extreme internet celebrities to telegram and alternative social media. *European Journal of Communication*. https://doi.org/10.1177%2F0267323120922066
[17] Stern, J. (2021) Social Media algorithms rule how we see the world. Last accessed: August 19th 2021. Sourced from: https://www.wsj.com/articles/social-media-algorithms-rule-how-we-see-the-world-good-luck-trying-to-stop-them-11610884800
[18] Novacic, I. (2020). Censorship on Social Media? It's not what you think. Accessed: August 19th 2021. Sourced from: https://www.cbsnews.com/news/censorship-social-media-conservative-liberal-cbsn-originals/
[19] Hassan, S., & de Filippi, P. (2017). The expansion of Algorithmic governance: From code is law to law is code. *The Journal of Field Actions*, Special Issue 17, pp 88-90.
[20] During 2020 a small number of research projects (including: AISEC and EnnCore) whose primary focus was verifiability and explainability of AI systems were funded by the EPSRC. A key target for AISEC is to imbue law and policy into AI code and verify the presence, application and impact of that law in the decisions or actions of the resulting AI.

*be used compatibly with road traffic law*[21]. While presenting as a positive enabling statement ostensibly supportive of live autonomous vehicle testing on UK roads, it is the potential scope encapsulated in the words *compatibly with road traffic law* that may have led those developing autonomous cars to believe that broad testing is still not fully permitted. It is possible some, rightly or wrongly, understood the statement to mean that the autonomous system itself must be capable of adherence to traffic law relevant to the operational activities performed by that technology. Ongoing consultations have clarified and reinforced that autonomous driving systems actually *should be* capable of traffic law compliance, including the Monitoring and Control Tests as defined in the DoT's 2020 *Centre for Connected & Autonomous Vehicles* Call for Evidence on *Safe Use of Automated Lane Keeping System* (ALKS)[22]. The primary monitoring test described at Section 5.1.2 calls for compliance with all road rules relevant to the dynamic driving task being performed by the active autonomous system. For example, this would mean a vehicle with ALKS and LCA should not cross double centre lines where the line closest to the vehicle is solid[23].

## 1.3. Advanced Driver Assistance Systems

The most significant advances for enabling autonomous vehicles have been the *advanced driver assistance systems* (ADAS). ADAS are technologies that improve or automate a function of the driving process previously performed by the human driver. The taxonomy used to classify and describe ADAS is that of the *Society for Automotive Engineers* (SAE) Automated Driving Levels (SAE-ADL) [28, 29]. The six-level SAE-ADL taxonomy shown in Figure 2 describe ADAS driving automation based on the technology's ability to subsume and automate the driving task. Levels 0-3 are differentiated by who controls two factors: (1) vehicle motion; and (2) the object event detection and response (OEDR) activities - which the SAE-ADL refer to collectively as the *dynamic driving task* (DDT). Levels 4 & 5 are differentiated based on whether the automation feature is capable of automation unrestricted by conditions imposed on or arising out of *geographic*, *road*, *environment*, *traffic*, *speed* or other limitations - which the SAE-ADL refer to as the *operational design domain* (ODD). Level 4 allows for automation constrained by one or more of these limitations at which point the driver must, or has the option to, take control of the vehicle. Level 5 allows for complete autonomous system independence. Table 1 provides a list of common ADAS and describes their function and classification based on SAE-ADL.

[21] Executive Summary, Findings, Point 9; https://assets.publishing.service.gov.uk/government/uploads/system/uploads/attachment_data/file/401562/pathway-driverless-cars-summary.pdf
[22]https://assets.publishing.service.gov.uk/government/uploads/system/uploads/attachment_data/file/980644/Safe-Use-of-Automated-Lane-Keeping-System-ALKS-Call-for-Evidence-FINAL-accessible.pdf
[23] The Highway Code, Rule 129 - which results by operation of *The Road Traffic Act* 1988 s36 and *The Traffic Signs Regulations and General Directions* 2002 s10 & 26.

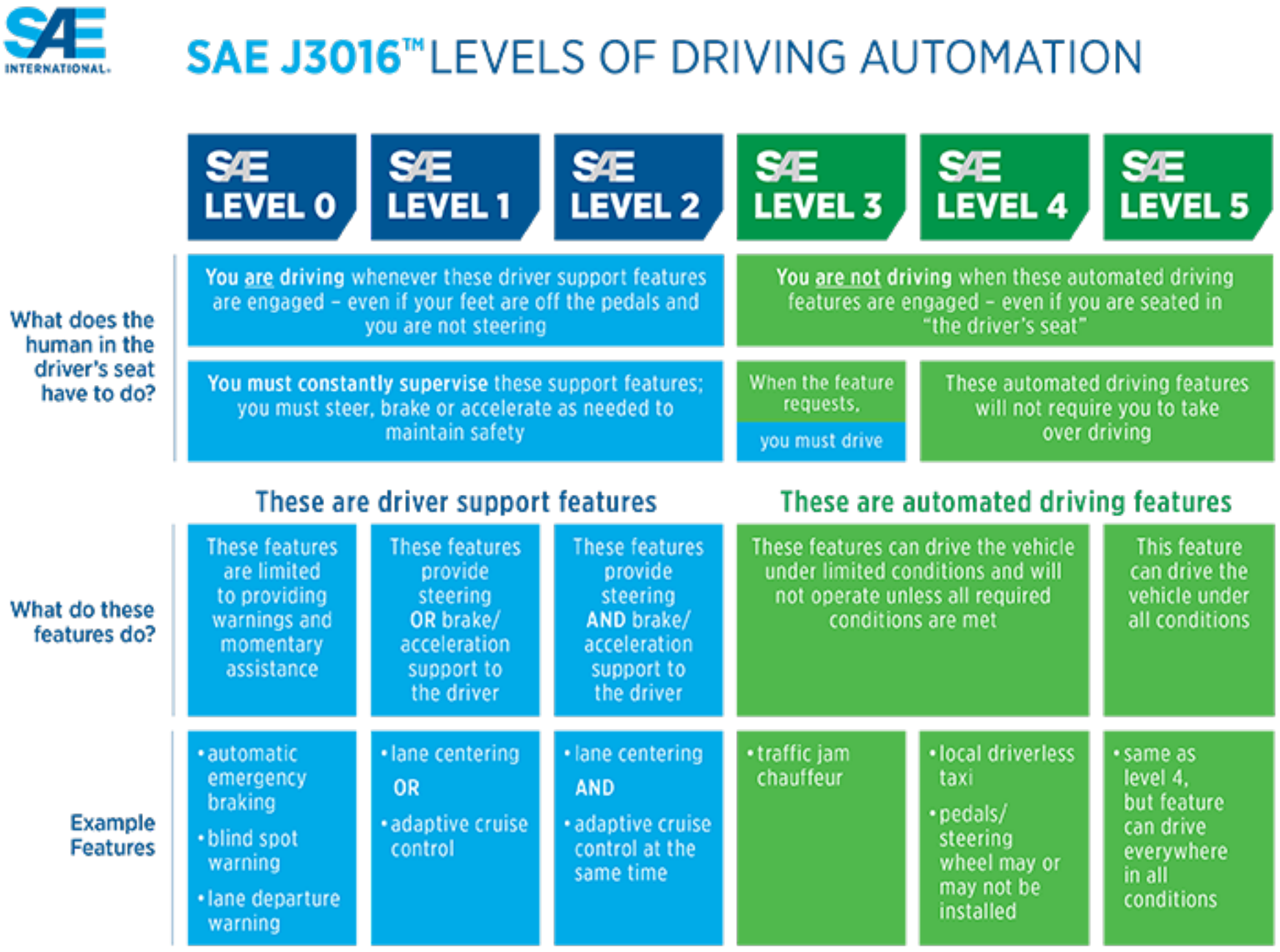

*Figure 2: SAE Six Levels of Driving Automation*

*Table 1: ADAS Technologies*

| ADAS | | DESCRIPTION | SAE Level |
|---|---|---|---|
| ABS | Anti-lock Braking System | Avoids wheel lock-up and loss of control during full-force braking. Uses wheel speed sensors in a feedback system to regulate brake pressure on individual wheels as they begin to skid. | 1 |
| ACC | Adaptive Cruise Control | Maintains desired speed while also ensuring a safe distance between the vehicle and the vehicle ahead. Uses mid- and long-range radar and forward-looking cameras to detect vehicles ahead and a feedback system to control braking and engine acceleration. | 1 |
| ANS | Automotive Navigation System | Provides location and turn-by-turn directions. Uses digital mapping tools and global positioning (GPS) signals. | 0 |
| AP | Automated Parking | Capable of identifying suitable parking spaces and taking control of some or all of the functions to set the gear selector, steer, accelerate and brake in order to park the vehicle. Uses radar-based parking sensors and cameras and control mechanisms for braking, steering and acceleration (where supported). | 1-2 |
| BSD | Blind Spot Detection | Provides visual and/or audible warnings when another vehicle is located in or crosses into the blind spot of your vehicle. Uses short-range radar sensors. | 0 |
| BUC | Back Up Camera | Automatically engages when the driver selects the reverse gear. Uses rear-facing camera. | 0 |
| DDD | Driver Drowsiness Detection | Provides audible and sometimes tactile alerts in situations where an algorithm believes the driver may be drowsy or otherwise inattentive. Uses motion sensors, internal driver-facing cameras and data on recent tactile inputs such as steering or indicator use. | 0 |
| ESC | Electronic Stability Control | Provides assistance to maintain vehicle stability to avoid oversteer and understeer through selective braking on a per-wheel basis. Uses accelerometers and gyroscopic sensors, wheel speed sensors, driver input sensors and information regarding engine rpm and torque. | 1 |
| FCM | Forward Collision Mitigation | Also known as Automated Emergency Braking System (AEBS), FCM is an upgrade on FCW that identifies situations where the driver has not reacted and autonomously applies the brakes. Uses the mid-range radar and forward-facing cameras and control mechanisms for braking. | 1 |
| FCW | Forward Collision Warning | Uses audible and visual warnings to alert the driver to an imminent frontal collision. Uses mid-range radar and forward-facing cameras. | 0 |
| HUD | Heads Up Display | Reduces the need for looking away from the road, displays essential information such as current speed, posted speed limit, current gear and turn-by-turn directions on windshield in front of driver. Uses input from multiple systems including speed sensors, TSR and ANS. | 0 |
| LCA | Lane Change Assistant | Performs autonomous lane changes either to ensure the vehicle is in the correct lane for a highway interchange or to exit a motorway in support of a course set in the navigation system, or as a result of commanded input from the driver. Uses | 2-3 |
| | | input from the LKC, PS, ACC, ANS and input from the cameras, lidar and ultrasonics to ensure a lane change is only performed when it is safe to do so. | |
| LDW | Lane Departure Warning | Provides audible and sometimes tactile warning if the driver accidentally wanders across the boundary of the current lane. Uses the forward-facing camera to detect the position of lane markings. | 0 |
| LKC | Lane Keeping and Centring | An extension of LDW that can prevent unintentional lane departure and return the vehicle to the centre of the current lane. Uses the forward-facing camera and a feedback system to apply suitable control inputs to the steering wheel. | 1 |
| NV | Night Vision | Uses fixed light infra-red (FLIR) thermal imaging cameras to enable vision of obstacles in very low light conditions. | 0 |
| PS | Parking Sensors | Provide audible alerts to assist in avoiding obstacles at very low speeds. Uses ultrasonic sensors positioned on the front and rear bumpers. | 0 |
| TCS | Traction Control System | Prevents the wheels from slipping by cutting torque and keeping the vehicle stable. Uses ESC sensor system. | 1 |
| TSR | Traffic Sign Recognition | Identifies and alerts the driver to the current speed and other posted road rules. Uses forward-facing cameras attached to the windshield and often integrate with information from ANS. | 0 |
| TJA | Traffic Jam Assistant | Subsumes the driving task in low-speed, high traffic congestion situations such as a traffic jam on a busy motorway. Uses cameras, lidar and ultrasonic sensors to identify the vehicle's in-lane position and that of other proximal vehicles, and is able to maintain distance, lane, and perform the entire task of low-speed stop-start driving. | 3 |

# 2. (De)Constructing simple representations of laws

Use of *Boolean Algebra* (Boole, 1847) for expressing the final structure of law or legal rules after thorough analysis, including rules of precedence, is not new[24]. More recently this approach has been applied in *Temporal* and *Boolean Logic* to model traffic law and road rules[25]. It's application in this work is intended to provide the basis for aiding AI developers and decision scientists in their efforts to develop machine learning (ML), neural networks (NN) and other forms of AI.

The process of deconstructing law and regulation has previously been demonstrated for property law and conveyancing processes[26], and occurs through investigation of the underlying structure of the law or regulation infers the inherent intention and flow; identifying the key points, actors, processes and chronology of operations and representing them by application of Boolean algebra and logic. For example, if we were examining the requirements of Section 5(1) of *The Motor Vehicles (Wearing of Seat Belts) Regulations* 1993 (MVWSBR) shown in Figure 3, which are already the underlying basis for Road Rule 99 shown in Figure 4, we see that: (a) the actors are adults and persons over the age of 14 years; (b) that they are driving or riding in a motor vehicle; and (c) they must wear a seat belt; (d) where that seat belt has been fitted or is available. Table 2 provides the resulting structured English pseudocode meeting the requirements of MVWSBR s5(1) as expressed by Road Rule 99.

[24] Kort, F. (1963). Simultaneous equations and Boolean algebra in the analysis of judicial decisions. *Law and Contemporary Problems*, 28(1), pp 143-163; and, Allen, L. & Caldwell, M. (1963). Modern logic and judicial decision making: A sketch of one view. *Law and Contemporary Problems*, 28(1), pp 213-270.

[25] Prakken, H. (2017). On the problem of making autonomous vehicles conform to traffic law. *Artificial Intelligence and Law*, 25, pp 341-363; and, Alves, G., Dennis, L., & Fisher, M. (2020). Formalisation and Implementation of Road Junction Rules on Autonomous Vehicle Modelling as an Agent. *Springer Nature*, Vol 12232. https://doi.org/10.1007/978-3-030-54994-7_16.

[26] McLachlan, S., Kyrimi, E., Dube, K., Fenton, N., & Webley, L. (2021). Lawmaps: Enabling Legal AI development through Visualisation of the Implicit Structure of Legislation and Lawyerly Process. *Manuscript accepted for publication in the J. of Artificial Intelligence and Law.* https://arxiv.org/pdf/2011.00586

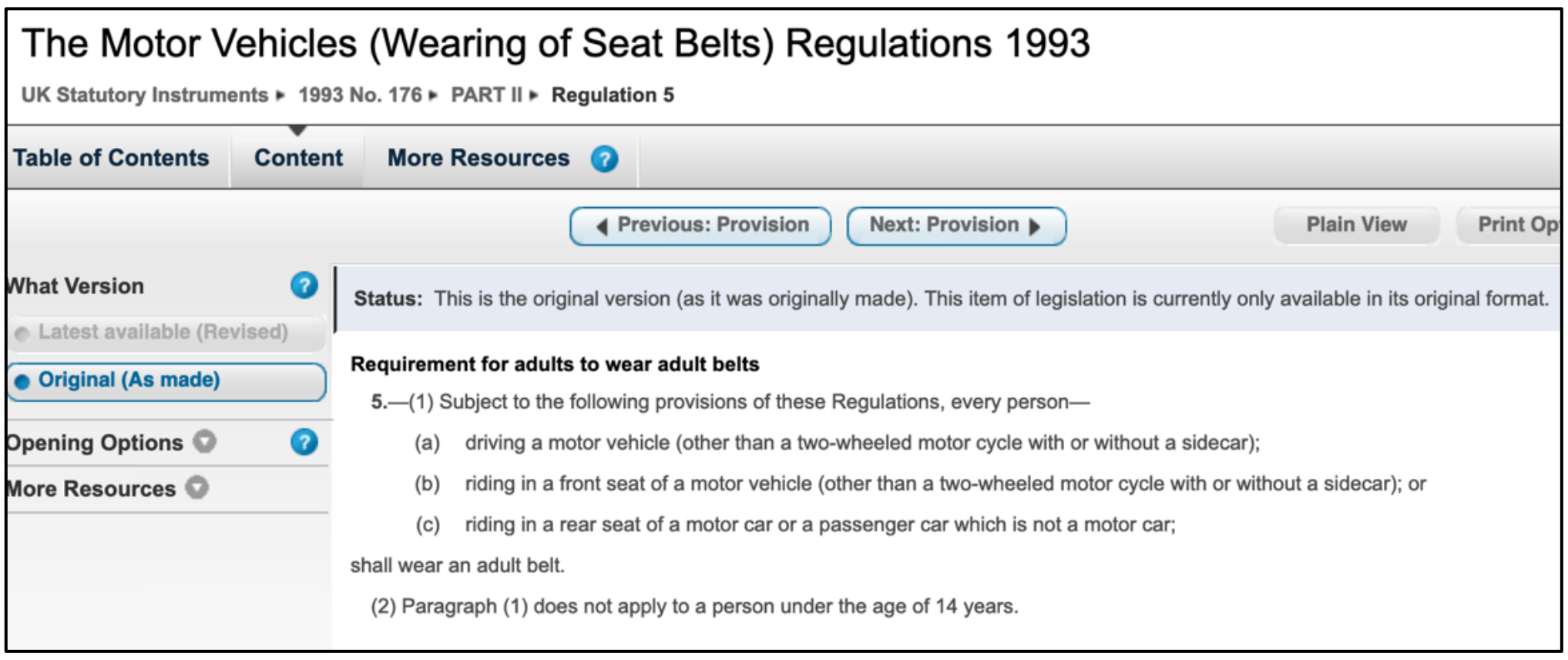
The Motor Vehicles (Wearing of Seat Belts) Regulations 1993

UK Statutory Instruments ▸ 1993 No. 176 ▸ PART II ▸ Regulation 5

Status: This is the original version (as it was originally made). This item of legislation is currently only available in its original format.

**Requirement for adults to wear adult belts**

5.—(1) Subject to the following provisions of these Regulations, every person—

(a) driving a motor vehicle (other than a two-wheeled motor cycle with or without a sidecar);

(b) riding in a front seat of a motor vehicle (other than a two-wheeled motor cycle with or without a sidecar); or

(c) riding in a rear seat of a motor car or a passenger car which is not a motor car;

shall wear an adult belt.

(2) Paragraph (1) does not apply to a person under the age of 14 years.

*Figure 3: MVWSBR s5(1)*

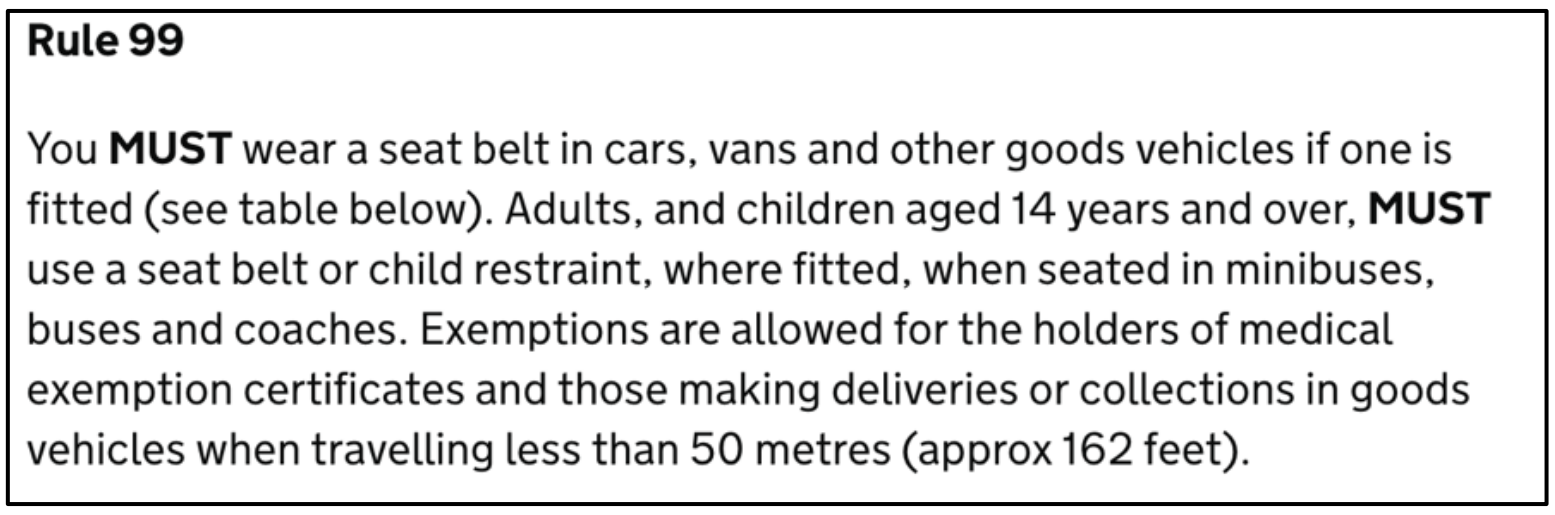
**Rule 99**

You **MUST** wear a seat belt in cars, vans and other goods vehicles if one is fitted (see table below). Adults, and children aged 14 years and over, **MUST** use a seat belt or child restraint, where fitted, when seated in minibuses, buses and coaches. Exemptions are allowed for the holders of medical exemption certificates and those making deliveries or collections in goods vehicles when travelling less than 50 metres (approx 162 feet).

*Figure 4: Road Rule 99*

*Table 2: Structured English logic for MVWSBR s5(1) as represented in Road Rule 99.*

```
IF:
    [A] Vehicle occupant is:
            a. An adult; or,
            b. A minor over:
                   i. 14 years of age; or,
                   ii. 1.35 metres in height.

EXCEPT:
    [C] Where seat belt is not fitted or available;

THEN:
    [X] Seat belt cannot be worn.

ELSE:
    [Y] Seat belt MUST be worn.
```

Boolean Algebra is well-established, mathematically sound, complete, and utilizes the following operators: *conjunction* (×, ∧), *disjunction* (+, ∨), and *negation* (~, ¬). This algebra is simple and is sufficient to express most legal rules in a way that most people would understand. Using Boolean Algebra, the Boolean logic and Boolean equations of Rule 99 would be expressed as shown in Table 3.

*Table 3: Boolean Logic and Boolean Equations for MVWSBR s5(1) as represented in the structured English*

| **Boolean Logic** | **Boolean Equations** |
|---|---|
| IF A and C<br>THEN X<br>ELSE IF A and NOT C<br>THEN Y | X = A × C<br>Y = A × ~C<br><br>where A = a + b |

These logic models simplify creation of machine interpretable state transition visualisations known as Lawmaps[27]; giving Lawmaps a sound formal basis. Lawmaps are an intelligent, clear and concise but democratised tool for visualising the structure and flow of law and lawyerly processes[28]. Lawmaps provide a visual primer of the potential paths and decision points that result from review of a particular law, regulatory rule or legal process. Exemplar Lawmaps are provided for Road Rules in the practical evaluation section of this work. We believe that construction of these logic models and their representation as Lawmaps is within the ability of legal practitioners, and will enable and expedite creation of legal software decision-making support tools, law-adherent AI, and validation of the consideration of and adherence to the rules of law in the decision-making processes and output of the tools and AI once created.

# 3. Method of Evaluation

This section presents the method used for a practical evaluation and assessment of whether current model vehicles with SAE Level 2 and 3 vehicle ADAS and autonomous systems are capable of adhering to the United Kingdom's (UK) traffic regulations as (de)constructed using the methods described in the previous section.

## 3.1. Process Used in Evaluation Experiments

In the method used for these evaluation experiments the abilities of three vehicles are assessed against the requirements of twenty-three UK Road Rules. Figure 5 presents the five step process followed in conducting this evaluation. Steps 1-3 are the *representation* steps that analyse the text of law and embody it in text more approachable to those from domains outside of law. Steps 4-5 focus on identifying and evaluating the relevant technology solutions employed by vehicle manufacturers. *Step 1* reviews, analyses and describes the rule of law under evaluation. This can include relevant legislation, policy and resulting regulations, byelaws or rules. In this step we provide a plain language description of the rule of law that we term a *structured English rule*, and compose the **rule description** section for each practical evaluation experiment. *Step 2* draws on the structured English rule and represents it in the form of Boolean logic and as a Boolean algorithm. *Step 3* visualises the process described by the structured English rule in the form of a Lawmap. This step also validates the visualisation by ensuring that a solution path can be identified through the Lawmap that is consistent with each Boolean algorithm. *Step 4* reviews the manual and other publicly available documentation to identify and describe the Manufacturer's technology responses relevant to the circumstances and requirements of this particular rule. During this step the **technology response** section is prepared for each practical evaluation experiment. The final step in the process, *Step 5*, evaluates and reports on the operational response of the technology in simulated and real-world conditions. This step seeks whether the vehicle's technology responds consistent to one (or more, as needed) of the required paths through the

[27] Ibid.
[28] Ibid.

Lawmap and the overall results are described in the **experimental findings** section for each practical evaluation experiment.

The evaluation experiments on the three vehicles were conducted over a period of three weeks along UK roads within the Greater London region. For each Road Rule the requirements and the vehicle systems and abilities relevant to that rule are described. A road test is also conducted to evaluate whether, and to what degree, each vehicle is capable of adhering to that Road Rule. Several examples of Boolean logic and Lawmaps are also provided to demonstrate how those processes were used in the conduct of this evaluation. The next two sections describe the three vehicles used in the evaluation experiments and the traffic light rating (RAG).

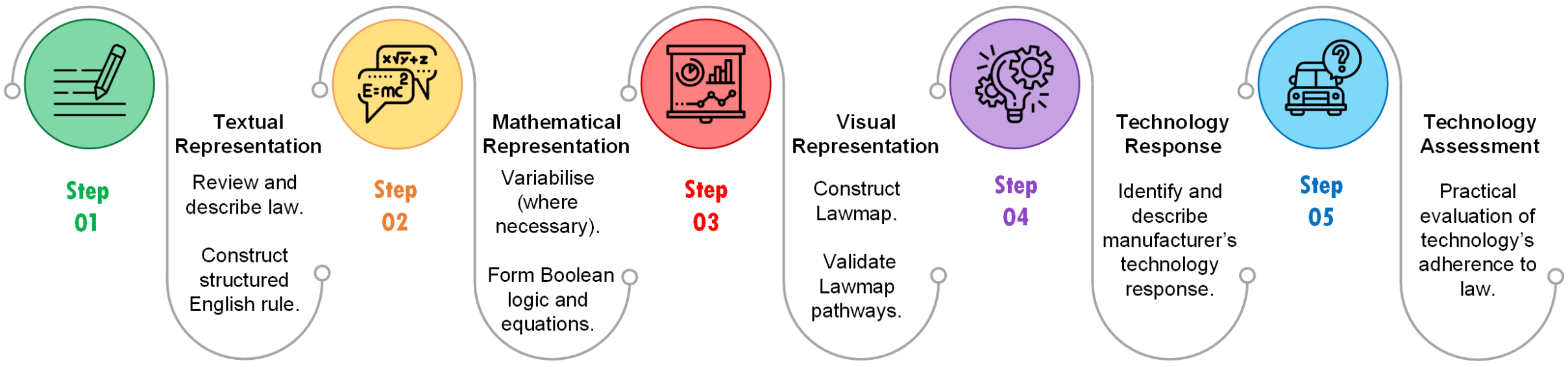

*Figure 5: Process for (de)constructing and representing law and evaluating smart vehicle technologies*

## 3.2. The Vehicles

Three vehicles were used in the testing reported in this work:

**2018 Opel/Vauxhall Insignia Grand Sport Elite Nav[29]**

The Vauxhall came equipped with ***Driver Assistance Package Four*** including *adaptive cruise control*, *forward collision alert* with *automatic braking*, *following distance indicator*, *lane departure warning*, *automatic lane keeping assistant*, *traffic sign recognition*, *blind spot detection* and a *360º camera system*. The vehicle came with a *heads up display* and an expanded feature stereo *head unit* that included 4G internet connectivity enabling live *traffic incident and congestion alerts*. This vehicle was also equipped with an advanced self-steering *automated parking assistant* solution. Between 2018 and 2020 this vehicle and specification was also badge engineered[30] as the Holden Calais-V in Australia and New Zealand and the Buick Regal Sportback and TourX in North America.

**2019 Mitsubishi Shogun Sport 4 Auto**

The Mitsubishi was supplied with a ***driver assist package*** including *blind spot warning*, *adaptive cruise control* with *forward collision alert*, *automatic braking* from a system called *forward collision mitigation*, and a *360º camera system*. This was the only vehicle to include *mis-acceleration detection and mitigation* - a system that identifies and prevents acceleration of the vehicle from a stationary position that would result in an impact with another stationary object, such as when Drive is unintentionally selected instead of Reverse when the vehicle is parked facing a solid wall in a parking garage. It was the only vehicle that did not include *automated lane keeping assist*. Since 2016 this vehicle and specification has also been sold by Mitsubishi under the model names Pajero Sport in Australia and New Zealand and Montero Sport in North America.

[29] Manufacturer Specification Brochure: https://www.dsg-vauxhall.co.uk/uploads/documents/insignia-pg.pdf
[30] *Badge engineering* is a common form of automotive market segmentation that describes a situation where multiple vehicle brands sell ostensibly the same vehicle with little or no actual engineering differences beyond the application of new badges, branding and logos both physically on the vehicle and in the software that operates the instrument cluster and multimedia user interfaces.

**2020 BMW 7 Series 740Li**

This luxury sedan was supplied with the *Technology Plus Pack* and ***Driver Assistant Professional*** (DAP) which BMW say make highway driving ***more autonomous***[31]. DAP provides *active navigation guidance*, *lane keeping assist*, *lane change assistant*, *emergency lane assistant*, *adaptive cruise control* with *automated braking*, *adaptive distance control*, *no passing indicator*, *forward collision avoidance*, *side and rear collision avoidance*, *pedestrian detection*, *lane departure* and *blind spot warning*, *rear cross traffic alert*, and a *traffic jam assistant* that subsumes the driving task in very low speed congested traffic. The vehicle also monitors the driver and provides audible and visual warnings if the driver's attention is diverted from the driving task, or when the drivers hands have been away from the steering wheel for too long, and can bring the car to a complete stop on the shoulder or emergency lane, where present, should the driver become unresponsive such as might occur in a cardiac health emergency.

## 3.3. Traffic Light (RAG) Rating

For each Road Rule that is evaluated, the rule description and test result a visual outcome indicator based on traffic lights (also known as a RAG rating) will also be provided. Table 3 describes the traffic light images and provides a definition for each colour rating.

*Table 4: Traffic lighting images and descriptions*

| | | |
|---|---|---|
| 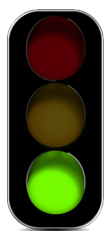 | 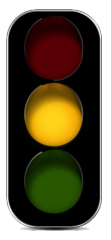 | 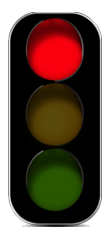 |
| **Green** | **Amber** | **Red** |
| ADAS solutions currently meet the all requirements of this Rule | Existing ADAS solutions could meet the requirements of this Rule with altered or extended software | Existing ADAS solutions cannot meet the requirements of this Rule without redesign and/or additional hardware |

# 4. Results

## 4.1 Evaluation Experiment

### (a) Rules 99-100

**Rule Description:** Road Rules 99-100 mandate all vehicle occupants use seat belts and/or approved child restraints. While it is possible in some distant future that vehicles may become so safe that occupant restraints are no longer necessary, for the foreseeable future these rules mandate the use of seat belts for adults and child restraints for children under 14 years of age or 1.35 meters in height.

[31] https://www.bmwblog.com/2020/05/27/bmw-driving-assistant-professional/

These rules exists as an amalgamation of: *The Road Traffic Act* 1988 s14 and 15, *The Motor Vehicles (Wearing of Seat Belts) Regulations* 1993, *The Motor Vehicles (Wearing of Seat Belts by Children in Front Seats) Regulations* 1993, and *The Motor Vehicles (Wearing of Seat Belts)(Amendment) Regulations* 2005 and 2006. Analysis of these legislation to create structured English logic resulted in three structured rules provided in Table 4 that together provide the sum of all requirements for Road Rules 99-100. It was necessary to variabilise the requirements of these structured rules before the Boolean equations in Table 6 could be constructed for them. The resulting Lawmap is provided in Figure 6 and a solution for each Boolean equation is provided in Table 7.

*Table 5: Structured English logic for Road Rules 99-100*

```
STRUCTURED RULE 1:
        IF:
            [A] Vehicle occupant is:
                    a. An adult; or,
                    b. A minor over:
                           i. 14 years of age; or,
                           ii. 1.35 metres in height.

        EXCEPT:
            [C] Where seat belt is not fitted or available;

        THEN:
            [X] Seat belt cannot be worn.

        ELSE:
            [Y] Seat belt MUST be worn.
```

```
STRUCTURED RULE 2:
        IF:
            [A] Vehicle occupant is a minor; and
            [B] Under 3 years of age.

        EXCEPT:
            [C] Where vehicle is a taxi; and,
                     a. Correct restraint is unavailable;
        THEN:
            [X] Minor may be unrestrained.

        ELSE:
            [Y] Correct child restraint MUST be used.
```

```
STRUCTURED RULE 3:
        IF:
            [A] Vehicle occupant is a minor;
            [B] 3 years of age or older; and,
                      a. Under:
                           i. 1.35 metres in height; or,
                           ii. 14 years of age.

        EXCEPT:
            [C] Where child restraint is:
                      a. unavailable;
                           i. In a licensed taxi or private hire vehicle; or,
                           ii. For reasons of unexpected necessity;
                                     a. over a short distance; or,
                           iii. If two occupied restraints prevent fitment of
        a third.

        THEN:
            [X] Adult restraint must be used.

        ELSE:
            [Y] Correct child restraint MUST be used;
                      a. Where:
                           i. Seat belts are fitted.
```

*Table 6: Variables and Boolean Equations for Road Rules 99-100*

| **Factual variables** | **Boolean Equations:** |
|---|---|
| q = (adult)<br>r = (child 14 yrs or over)<br>s = (child over 1.35 m tall)<br>t = (baby under 3)<br>u = (child 3 yrs or over)<br>v = (child under 14 yrs)<br>w = (under 1.35 m tall)<br><br>**Situation variables:**<br>p = (where child restraint is unavailable:<br>(i) taxi or<br>(ii) unexpected necessity or<br>(iii) two occupied restraints)<br>x = (where seat belts are fitted)<br>y = (where seat belt is neither fitted nor available)<br>z = (Where vehicle is a taxi and correct restraint is unavailable)<br><br>**Decision variables:**<br>A = Minor may be unrestrained.<br>B = Seat belt cannot be worn.<br>C = Adult restraint must be used.<br>D = Seat belt MUST be worn.<br>E = Correct child restraint MUST be used.<br>F = E ∧ x | *Structured Rule 1:*<br><br>$B = (q \lor (r \lor s)) \land y$<br>$D = (q \lor (r \lor s)) \land \neg y$<br><br>*Structured Rule 2:*<br><br>$A = t \land z$<br>$E = t \land \neg z$<br><br>*Structured Rule 3 :*<br><br>$C = (u \lor v \lor w) \land p$<br>$F = (u \lor v \lor w) \land \neg p$ |

*Table 7: Lawmap for Road Rules 99-100 showing path for each Boolean equation solution*

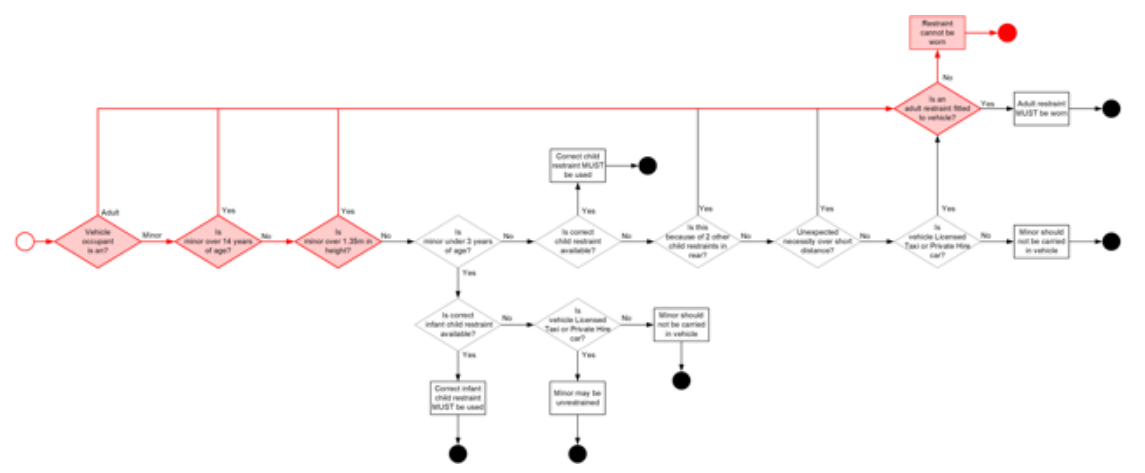

*Lawmap Solved for B*

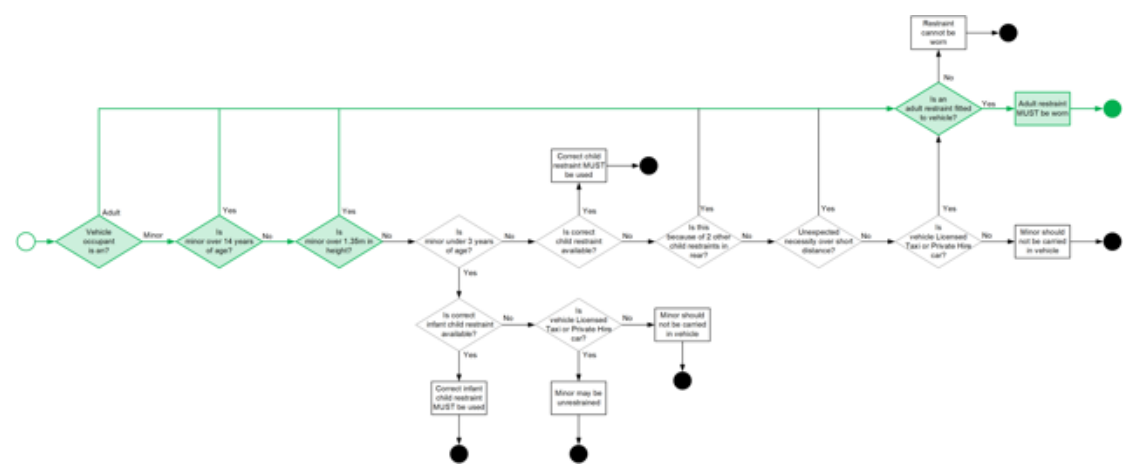

*Lawmap Solved for D*

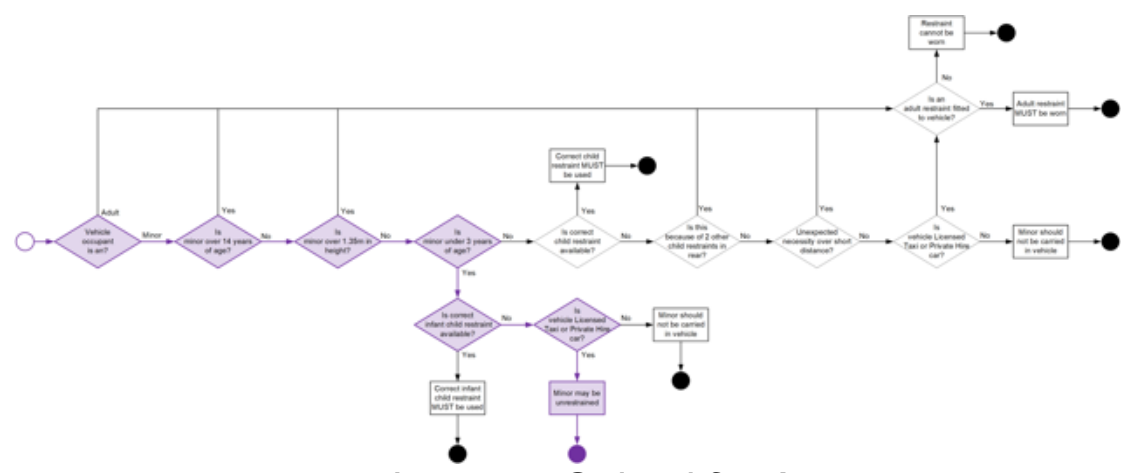

*Lawmap Solved for A*

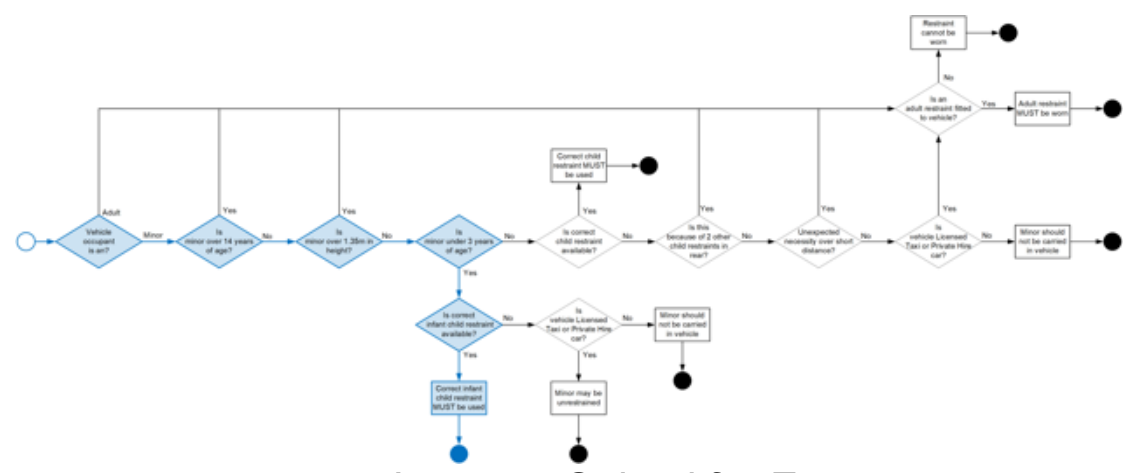

*Lawmap Solved for E*

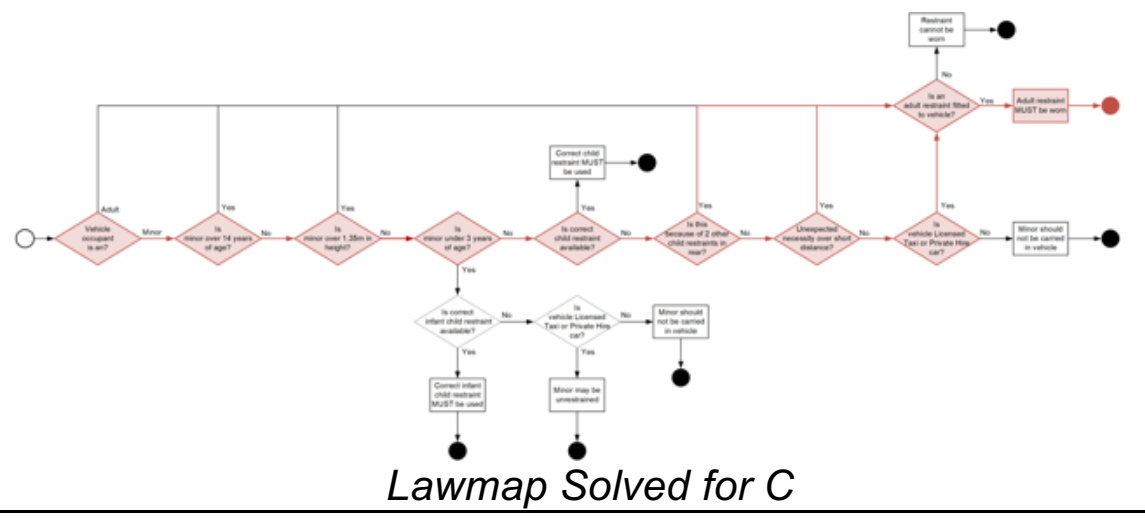

*Lawmap Solved for C*

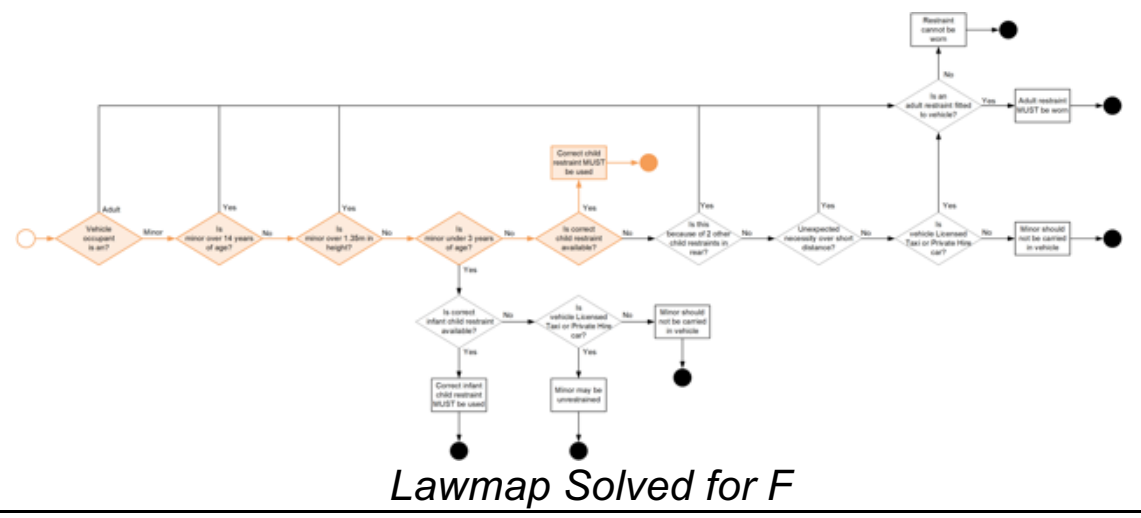

*Lawmap Solved for F*

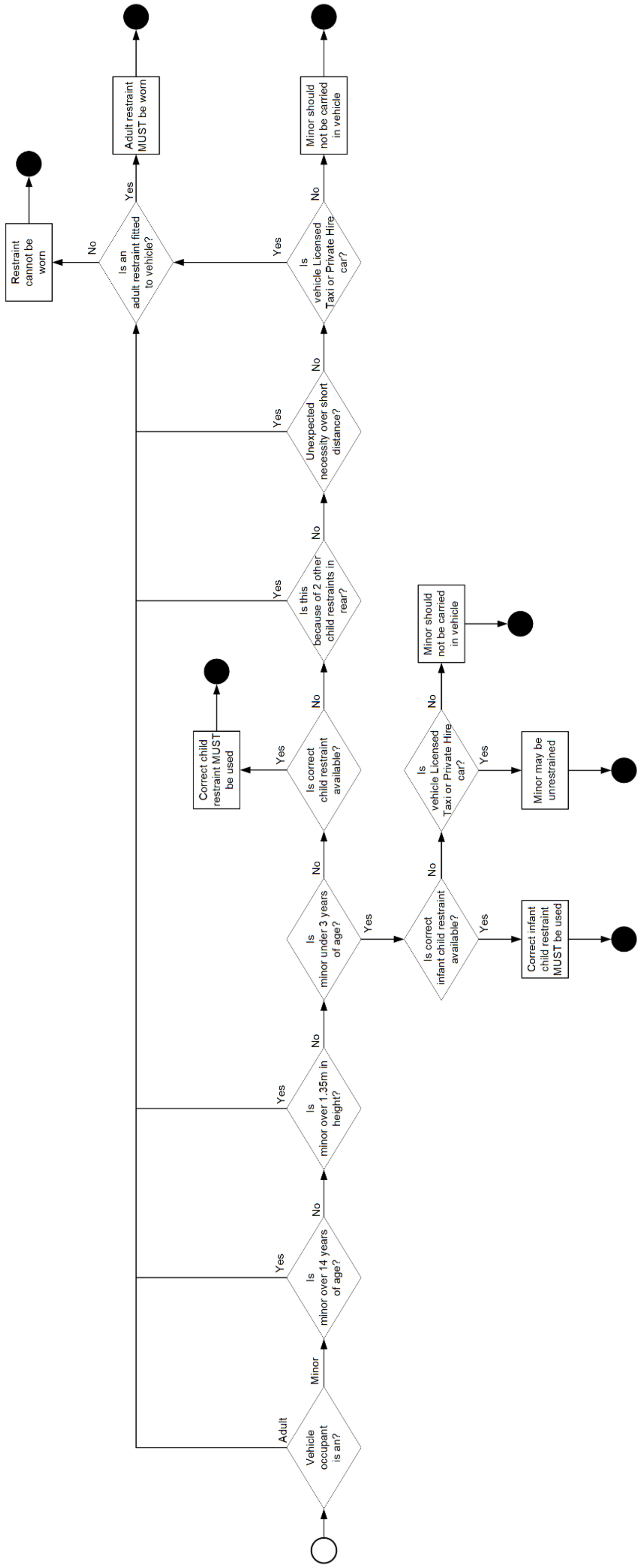

*Figure 6: Lawmap for Road Rules 99-100*

**Technology Response:** Using pressure sensors in some seats and circuit detection in the seat belt connection point, most current vehicles are able to indicate using visual and/or audible warnings either when a seat is occupied, or when a seat belt is not currently in use. All of the test vehicles provided a visual indicator showing all seats in the vehicle. On engine start all three vehicles illuminated indicators for the three rear seats irrespective of whether they were occupied or not. If at any point after driving off a rear seat passenger in the Vauxhall unplugged a seatbelt that was plugged in when the car had started to move, the indicator for that seat would illuminate red. The other vehicles initially only lit indicators for all three rear seats whether or not the seat was occupied or the seatbelt was plugged in. On reviewing and discussing notes from these tests, a BMW technician verified the presence of sensors in all three rear seat belt receivers, and circuit-tested them in an effort to identify why they were not indicating in the instrument cluster when a rear seat passenger unplugged a seat belt. A software update was received from BMW after initial testing was complete but prior to submission of this work that, as shown in the right-most image of Figure 7, enabled green illumination when a rear seat passenger plugged in their seatbelt, and red illumination and an alert if they later unplugged that seatbelt while the vehicle gear selector remained in Drive.

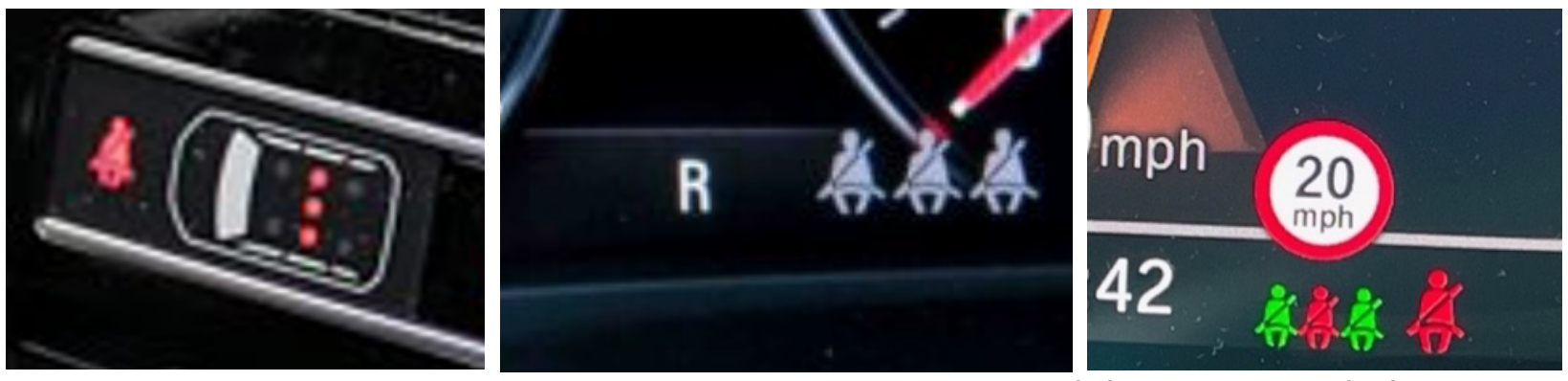

*Figure 7: Rear passenger seatbelt indicators on Mitsubishi (L), Vauxhall (M), and BMW (R)*

Bayesian Networks (BNs), also described as *causal probabilistic models* or *belief networks*, are *directed acyclic graphical* (DAG) models[32] in which the nodes represent variables and the arcs represent causal, probabilistic, or influential relationships between variables. BN models have previously been used to represent and evaluate legal arguments[33], to compute the probability for whether an adjudicated outcome was correctly or randomly decided[34], to calculate the probability for success in litigation[35], and to evaluate the likelihood of guilt given the presence and weighted credibility of particular evidence[36]. Drawing on the structured rules and Boolean equations to identify requirements that are represented by the *nodes* and relationships between those requirements which are denoted by the *arcs*, a BN for *Road Rules 99-100* was developed in AgenaRisk (https://www.agenarisk.com) and is shown in Figure

[32] J. Pearl, Probabilistic reasoning in intelligent systems: networks of plausible inference. Morgan Kaufmann Publishers Inc., 1988
[33] Neil, M., Fenton, N., Lagnado, D., & Gill, R. D. (2019). Modelling competing legal arguments using Bayesian model comparison and averaging. *Artificial intelligence and law*, *27*(4), 403-430.
[34] Guerra-Pujol, F. E. (2011). A Bayesian model of the litigation game. *Eur. J. Legal Stud.*, *4*, 204.
[35] McLachlan, S., Kyrimi, E., & Fenton, N.E. (2019). Public Authorities as Defendant: Using Bayesian Networks to determine the Likelihood of Success for Negligence claims in the wake of Oakden. Preprint available: https://arxiv.org/abs/2002.05664
[36] Fenton, N., Neil, M., Yet, B., & Lagnado, D. (2020). Analyzing the Simonshaven case using Bayesian networks. *Topics in cognitive science*, *12*(4), 1092-1114.

8. This model was validated through instantiation of the variables of each Boolean equation as *observations* on the model. Figure 9 demonstrates the solution when we observe all variables required for the scenario represented by the Boolean equation for C resulting from Structured Rule 3.

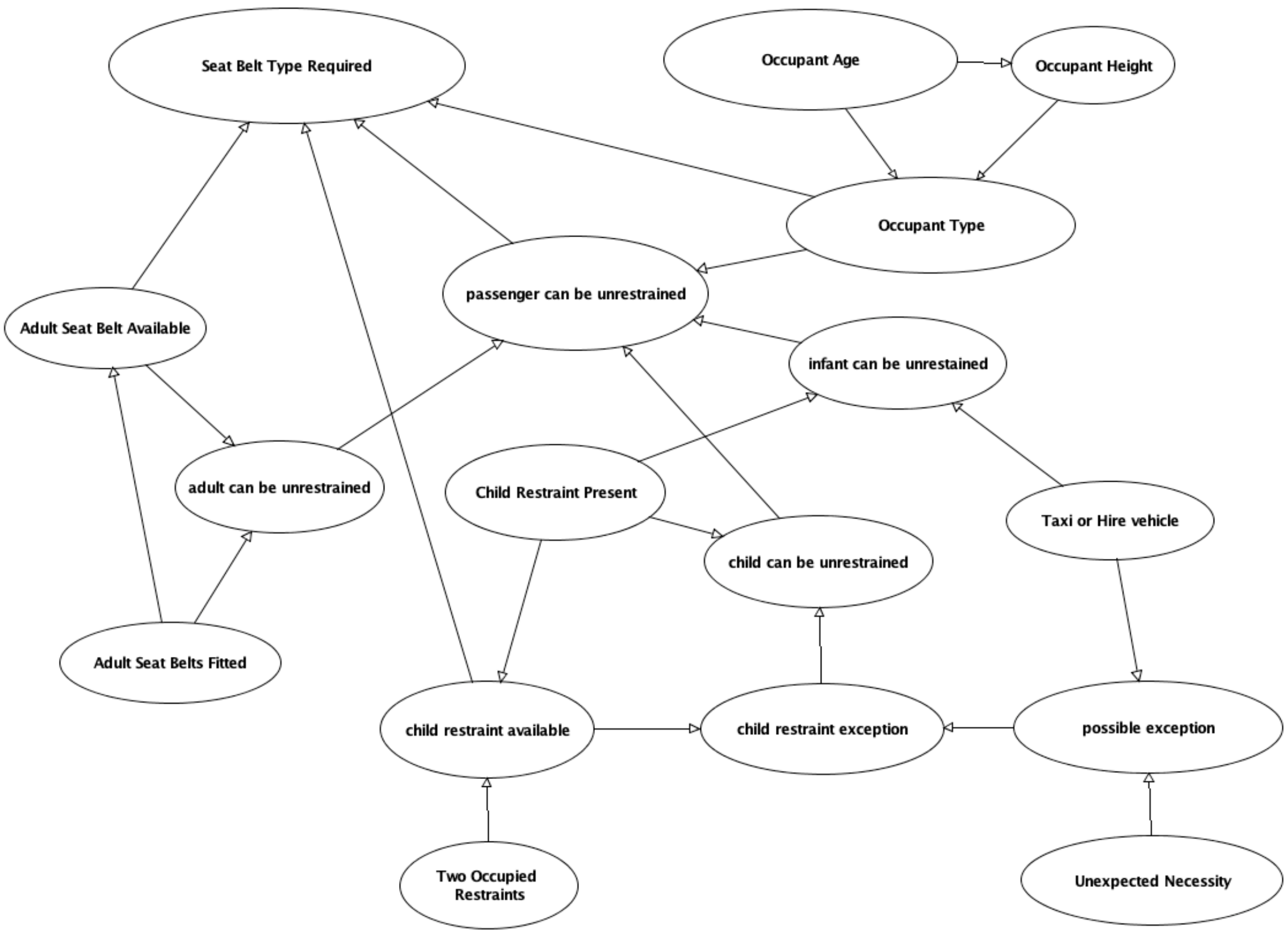

Figure 8: BN model for Road Rules 99-100

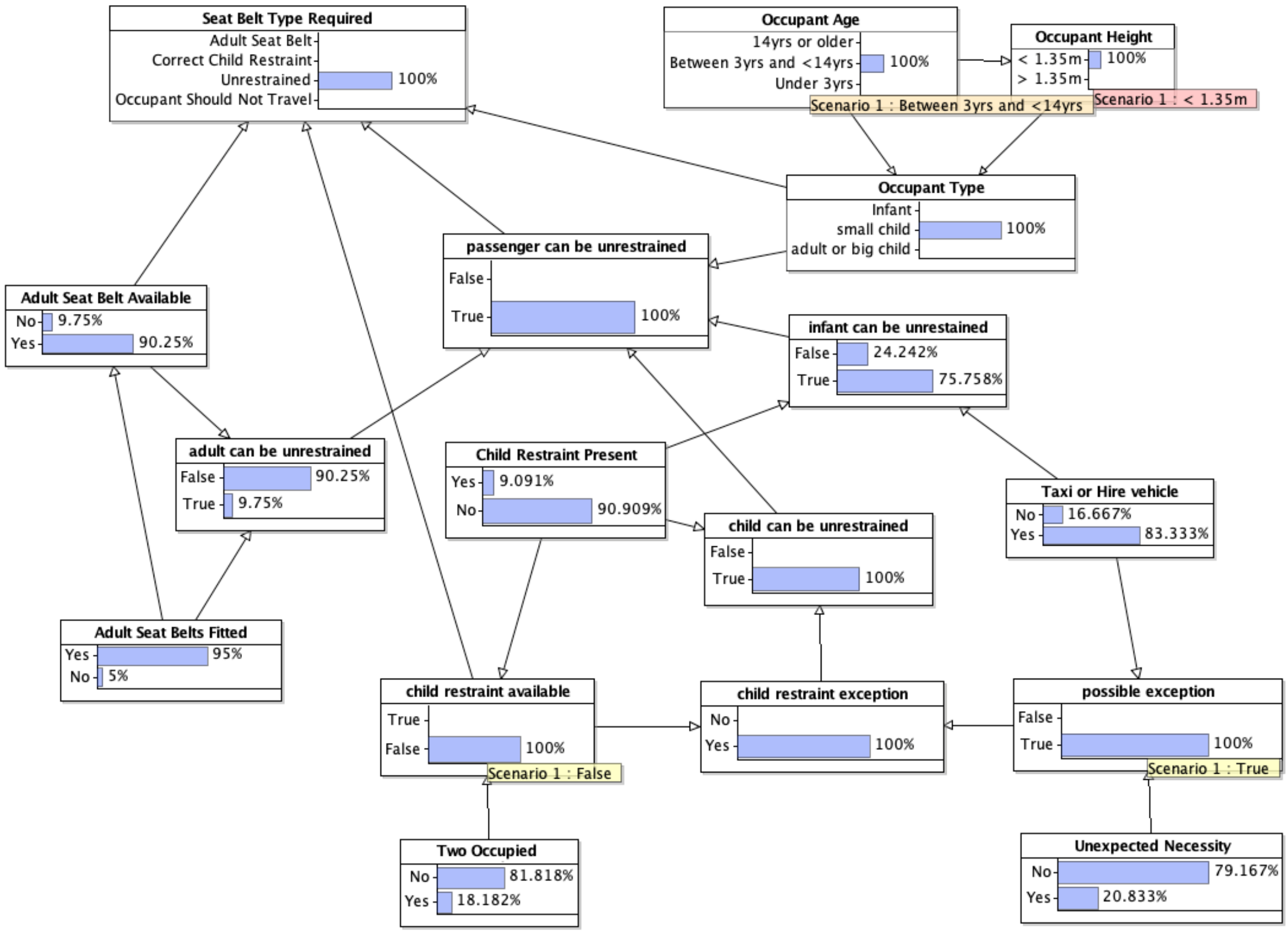

*Figure 9: BN model for scenario representing Boolean equation C*

**Experimental Findings:** The Vauxhall extinguished the indicator for any given seat when the seatbelt was plugged into the receptacle. All three vehicles used an audible warning if one of the front seat occupants unplugged their seatbelt while the vehicle was moving. Two (Vauxhall and BMW) also alerted when a rear seat passenger unplugged their seatbelt while the vehicle was in motion. However, none of the vehicles ceased operation of ADAS systems if any seatbelt was unplugged. Given that other ADAS and autonomous functions were capable of safely bringing the vehicle to a stop either in-lane or on a hard shoulder, it would be a simple matter for future vehicles or software updates to extend an alert to unrestrained occupants and, if that warning goes unheeded, to safely cease travel until occupants engage their seat belt. Future vehicles should also come installed with pressure sensors in all seats to ensure the driver or autonomous system is aware of the presence and seat location of all occupants. However, new technology would need to be able to measure the height or estimate the age of child passengers and determine whether a child restraint was required.

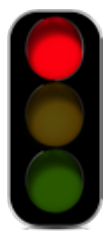

## (b) Rules 103-105

**Rule Description:** Rules 103-105[37] concern use of and attention to various signals and traffic directions. To: (a) use indicators and brake lights to signal other road users and pedestrians prior to and as intended changes to course, direction and speed occur[38]; (b) cancel the engaged signal after use[39]; (c) be attentive to for the indicator and brake light signals of other vehicles and proceed only when it is safe to do so[40]; (d) remain alert for situations where it is possible that the signal of another vehicle may be erroneous because it has not been cancelled[41]; and (e) obey the traffic directions of a range of authorised persons[42]. This final provision arises by operation of: *The Road Traffic Regulation Act* 1984 (RTRA) s28, *The Road Traffic Act* 1988 (RTA) s35, and *The Function of Traffic Wardens (Amendment) Order* 2002 (FTWAO). These laws collectively mandate compliance with the traffic directions of constables, traffic officers, traffic wardens, and school crossing patrols. Table 8 provides the structured English logic for the general operation of Road Rule 103, which results in the Boolean logic and equations in Table 9, and was then applied in development of the Lawmap shown in Figure 10. Table 10 and Figure 11 provide an example scenario for Road Rule 103 where the intention is to pull off and stop on the side of the road just after a side road.

[37] https://www.gov.uk/guidance/the-highway-code/general-rules-techniques-and-advice-for-all-drivers-and-riders-103-to-158
[38] Rule 103.

*Table 8: Structured English logic for the general operation of Road Rule 103*

```
IF:
  [A] When in control of a motor vehicle; and,
        [B] There is an intention to:
             a. Change course; or,
             b. Direction; or,
             c. Stop; or,
             d. Move off.

EXCEPT:
  [C] Where it would be misleading to signal at that time;

THEN:
  [X] Signalling should be delayed;
          a. until:
                i. Signalling would not be misleading;

ELSE:
  [Y] Other road users should be alerted by:
          a. Clear signals;
          b. Given in plenty of time.
```

*Table 9: Boolean Logic and Boolean Equations for Road Rule 103*

| **Boolean Logic** | **Boolean Equations** |
|---|---|
| IF (A and B)and C<br>THEN X<br>ELSE IF (A and B)and NOT C<br>THEN Y | X = (AxB) × C<br>Y = (AxB) × ~C |

**Technology Response:** Two test vehicles (Vauxhall and BMW) came equipped with camera-based systems affixed to the upper windscreen in the area above and forward of the rear-vision mirror. These were in addition to forward-facing ultrasonic sensors and, in the case of the BMW, radar. Both make claims in their driver manuals and online marketing materials regarding these vehicles ability to detect and identify speed limit signs, however BMW goes one step further and promotes its system as capable of recognising a range of road signs to warn drivers when they must yield (*stop* or *give way*), or even when they are entering a restricted one-way street[43].

[39] Ibid.
[40] Rule 104.
[41] Ibid.
[42] Rule 105.

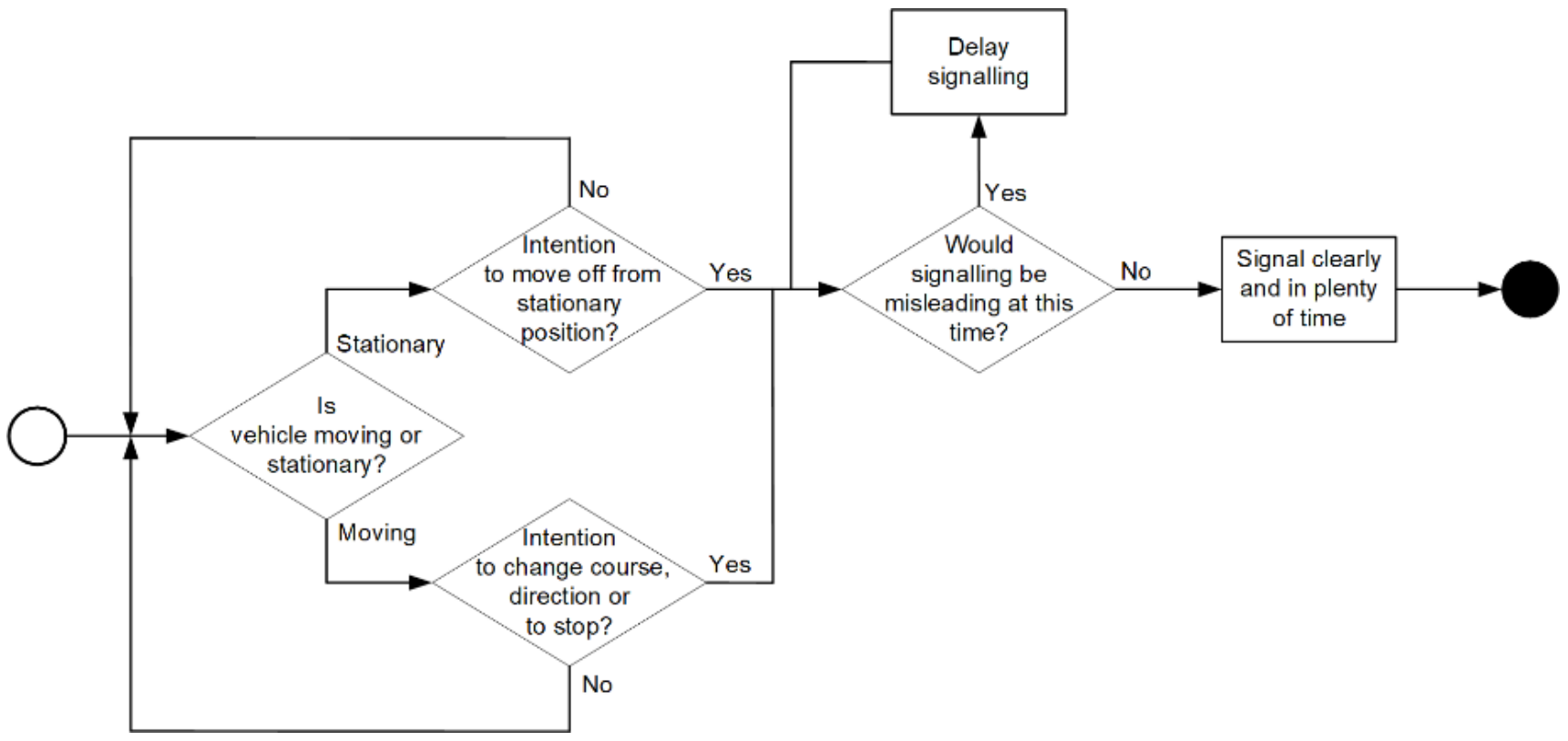

*Figure 10: Simple Lawmap for the general operation of Road Rule 103*

*Table 10: Structured English logic for an example scenario application of Road Rule 103*

```
IF:
  [A] When in control of a motor vehicle; and,
  [B] There is an intention to:
            a. Turn or exit the current road;
                i. Immediately after passing a side road on;
                ii. The same side as the intended turn.

EXCEPT:
  [C] Where it would be misleading to signal at that time;
            b. Because other drivers may believe it signals an intention to;
                i. Turn into the side road.

THEN:
  [X] Signalling should be delayed;
            a. Until:
                i. The vehicle has passed the side road.

ELSE:
  [Y] Other road users should be alerted by:
            a. Clear signals;
                i. Brake lights to warn when slowing down; and,
                ii. Indicators to warn of change in course;
            b. Given with sufficient time to;
                i. Adjust their own course and speed; and,
                ii. Avoid potential for an accident.
```

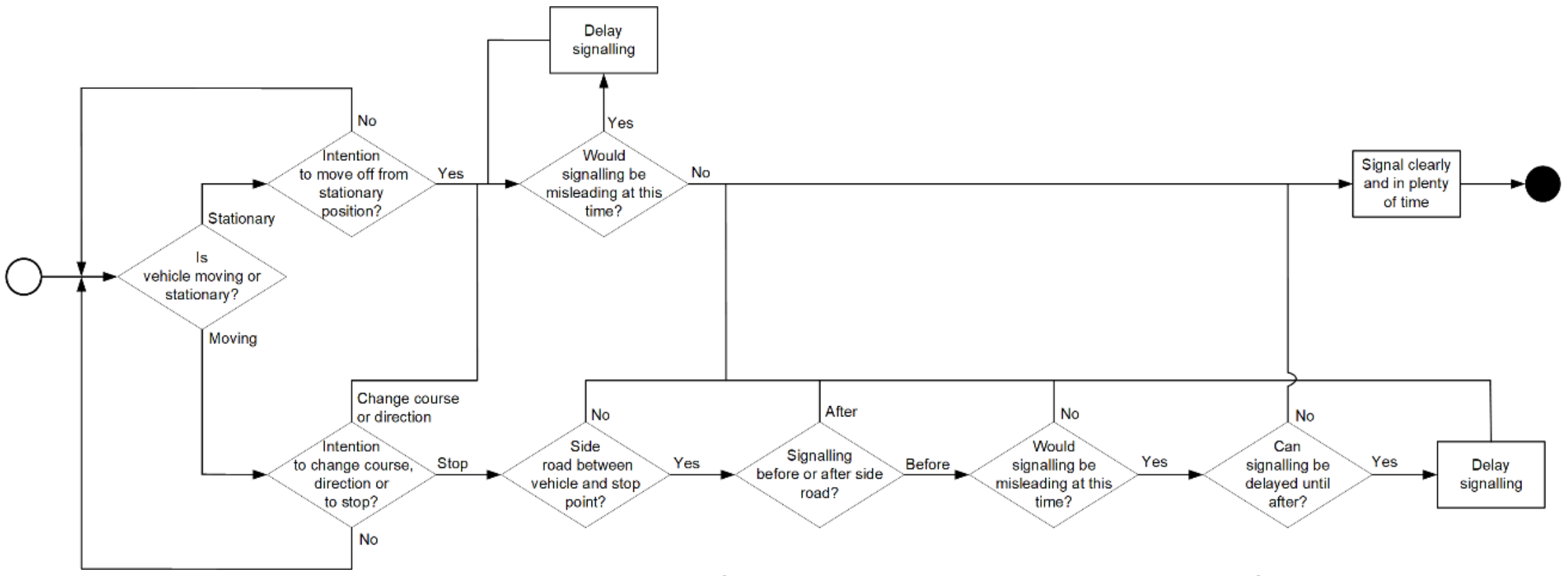

*Figure 11: Extended simple Lawmap for the example scenario operation of Road Rule 103*

---

[43] https://www.bmw.com/en/innovation/the-main-driver-assistance-systems.html#pwjt-road-sign-recognition

**Experimental Findings:** While none of the vehicles was able to identify when indicator or brake lights were illuminated on the vehicle they immediately followed, all were able to detect and react appropriately when that vehicle's forward velocity reduced, whether passively or because the brakes had been applied. In each case all three vehicles first alerted to the changed velocity and potential for collision and, if no immediate corrective action was taken, each applied the brakes and was capable of autonomously bringing the vehicle to a complete stop. The Vauxhall and BMW vehicles both identified speed limit signs and adjusted the displayed speed limit in both the instrument cluster and heads up display, as shown in Figure 12. However, on occasion both were prone to error. One example of these errors is shown in Figure 13 where the BMW vehicle on repeat trips past a 30mph sign on an urban lane consistently mis-read the value as 80mph. Other examples occurred where the speed limit sign was small, or where stickers on the rear of a delivery van or lorry (Figure 14) were incorrectly identified as a speed limit traffic sign. Our evaluation of each speed limit error suggested the vision-based value overrides the *known good* value provided by the map system[44], resulting in the mis-read speed being shown to the driver as the current speed limit.

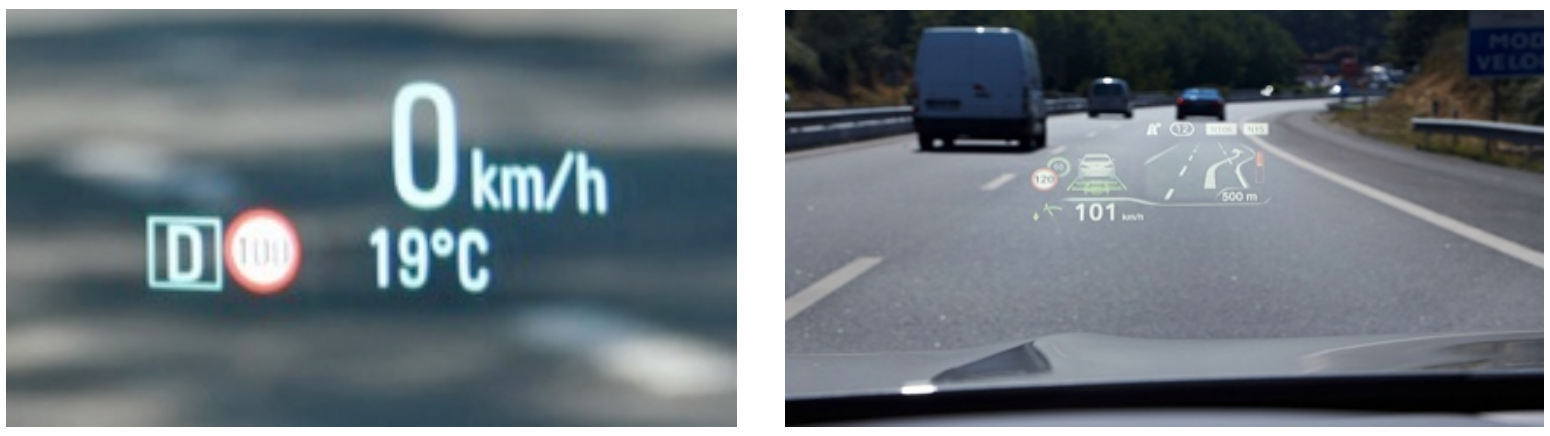

*Figure 12:Heads Up Display (HUD) showing Speed Limit Sign Detection. Vauxhall (L) and BMW (R)*

Of the three vehicles, only the BMW *Driver Assistant Professional* (DAP) was capable of and correctly identified fixed mounted *Give Way* and *Stop* signs. However, the BMW system failed to identify a *Stop* sign when it was held by a school crossing patrol guard who, in almost 50% of instances, it also incorrectly identified as a stationary small vehicle. Also, the BMW vehicle did not apply or perform the direction given of these signs, even when in full DAP *SAE Level 3* mode. Finally, none of the tested vehicles identified traffic lights, a capability now found in some *hardware three* (HW3) based Tesla Autopilot equipped vehicles.

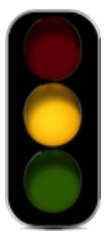

[44] We believe this was intended as a safety feature whereby the system would update and alert the driver to reduced speed limits, for example when indicated on temporary signs for roadworks.

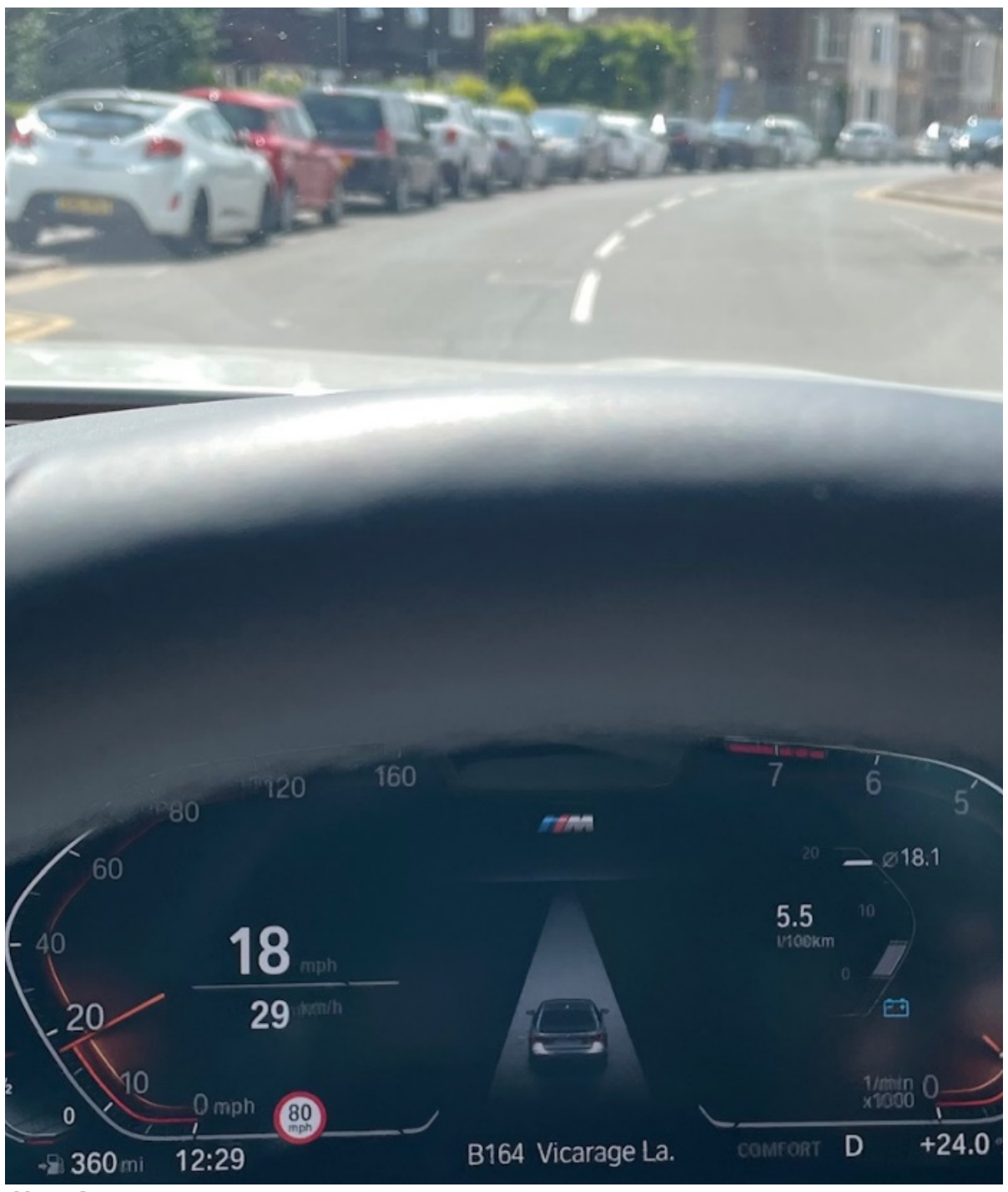

*Figure 13: Example of Traffic Sign recognition error indicating 80mph in a 30mph urban area (Vehicle: BMW)*

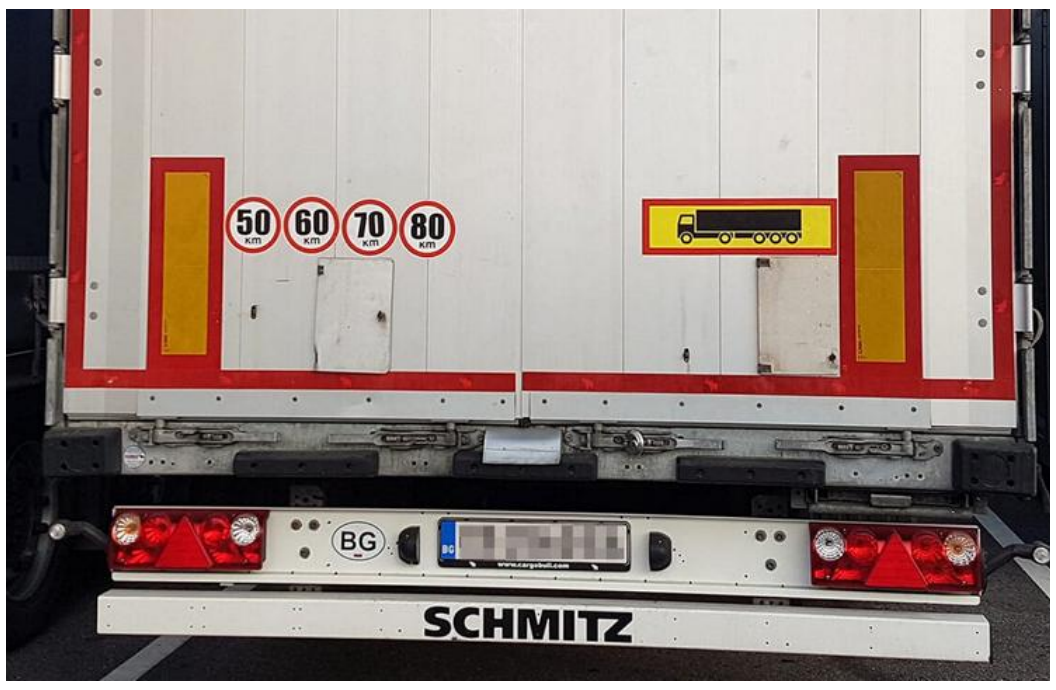

*Figure 14: Speed limit stickers on the rear of a Semi Truck Trailer*

## (c) Rule 113

**Rule Description:** Rule 113 requires all sidelights and rear registration plate lights are lit between sunrise and sunset, that headlights are used at night[45] except on a road which has lit street lighting[46], and when visibility is seriously reduced. This rule arises from *The Road Vehicles Lighting Regulations* 1989 s3, 24 and 25.

**Technology Response:** Like most current model vehicles, all three in this test were fitted with automatic headlights as standard. All three used photoelectric sensors to identify low light situations.

[45] Night, or the hours of darkness, are described in the Road Rules as the period between half an hour after sunset and half an hour before sunrise.
[46] The notes with Rule 113 identify these as roads that are generally restricted to a speed limit of 30mph unless otherwise specified.

**Experimental Findings:** Each vehicle correctly identified low light situations and, when triggered by the sensor, activated the headlights, tail lights and registration plate lights.

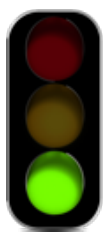

## (c) Rules 127-132

**Rule Description:** Rules 127-132 describe the standard for line markings and dividers used on UK roads. This includes the broken white line at the centre which a vehicle should not cross unless the road is clear and there is an intention to overtake or turn off; and which, when the line lengthens and gaps shorten, indicates a hazard ahead. It also includes where there are double white lines at the centre of the road that generally mean you *must not* straddle or cross the centre line; *unless* (a) the line closest to you is broken which indicates that overtaking is permissible; (b) it is safe to do so and you are passing a stationary vehicle or overtake a bicycle, horse or road maintenance vehicle; or (c) you are crossing to enter an adjoining property or side road.

**Technology Response:** Two vehicles (Vauxhall and BMW) included automated lane keeping and centring. Both rely on visual input from windscreen-mounted cameras to identify lane markings and have escalating levels of functionality from warning the driver when they are about to cross a lane boundary through to an ability to take corrective action by autonomously steering the vehicle back into the centre of the current lane. Only the BMW was also fitted with *lane change assistant* (LCA), an additional software system that uses the camera, lidar and blind spot detectors and receives input from the driver for an intended lane change. LCA scans if it is safe to proceed in the commanded direction and then autonomously performs the lane change operation. The Mitsubishi was not fitted with a lane keeping assistant feature and was unable to be tested against this rule.

**Experimental Findings:** The Vauxhall and BMW vehicles correctly identified and provided an audible alert when the vehicle was about to cross out of a lane. The BMW vehicle's next level of response was to perceptively vibrate the steering wheel without affecting driveability or the current vehicle heading. The final level for both was to autonomously steer the vehicle back into the centre of the current lane. When tested, both could be overridden by the driver through the application of slightly greater than normal force on the steering wheel in the driver's chosen direction. Testing of the BMW's LCA showed that it does not differentiate circumstances where there were double centre lines, nor whether the nearest line was unbroken. On a large four-lane (two-a-side) country road it was still willing to perform a lane change manoeuvre that would have resulted in crossing the double solid centre line.

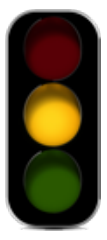

## (d) Rules 137-138

**Rule Description:** Rules 137-138 govern the use of different lanes on dual carriageways, including for overtaking and right-hand turns. Rule 137 provides that a vehicle may use the right-hand lane of a two-lane dual carriageway and Rule 138 extends this by allowing use of the middle or right lanes on a three-lane carriageway for overtaking or performing a right turn. However, both rules stipulate driving in the left-most lane generally, and where the vehicle has left the left-most lane to overtake, returning to the left lane when it is safe to do so. The structured English logic for Rules 137-138 is provided in Table 11, and the Boolean logic and equations are provided in Table 12.

*Table 11: Structured English logic for Rules 137-138*

```
IF:
  [A] Where a vehicle is on a two-lane or three-lane dual carriageway;

EXCEPT:
  [C] Where there is an intention to:
             a. Overtake; or,
             b. Turn right;

THEN:
  [X] The vehicle may:
             a. Use:
                 i. The right lane on a two-lane dual carriageway; or,
                 ii. The middle or right lane on a three-lane dual carriageway
             b. Until:
                 i. It is safe to move back into the left lane.

ELSE:
  [Y] The vehicle should stay in the left lane.
```

*Table 12: Boolean Logic and Boolean Equations for Road Rule 137-138*

| Boolean Logic | Boolean Equations |
|---|---|
| IF A and C<br>THEN X<br>ELSE IF A and NOT C<br>THEN Y | $X = A \times C$<br>$Y = A \times \sim C$ |

**Technology Response:** While both the Vauxhall and BMW were fitted with LDW and LKA, only the BMW also had LCA. BMW's LCA and LDW work together using the various ultrasonic, radar and camera sensors fitted to the vehicle to ensure that the vehicle's LCA only changes lanes where it is safe to do so[47].

**Experimental Findings:** BMW's current iteration of LCA operates in two modes: one autonomous and the other semi-autonomous. The *first* or autonomous mode occurs in full DAP with navigation enabled, and will perform some lane changes on main highways and freeways to ensure the vehicle is in the correct lane to exit or change highways as required by a pre-programmed course. The *second* or semi-autonomous mode performs lane changes only after commanded input from the human driver. In this second mode the driver is required to apply a small amount of pressure to the indicator stalk in the direction of the intended change, and the system commences scanning to ascertain whether it is safe before taking control of the direction of travel via steering inputs necessary to perform the requested lane change. Neither mode saw the vehicle autonomously return to the left-most lane as required by Rules 137-138 without additional commanded input on the indicator stalk.

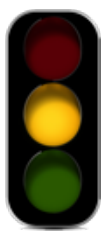

[47] https://www.bmw.com/en/innovation/the-main-driver-assistance-systems.html#pwjt-lane-change-assist

## (e) Rules 191-199

**Rule Description:** Rules 191-199[48] provide directions for approaching and negotiating pedestrian crossings. The UK has several types of pedestrian crossings defined as *controlled* and *uncontrolled*: differentiated by the presence or absence of traffic lights and how those traffic lights function. How crossings function is also altered by the presence of pedestrian havens, known as *central islands*. Uncontrolled crossings include standard zebra crossings, zebra crossings with central islands, and school warden crossings. There are four types of controlled crossing: (i) The *Pelican Crossing* is controlled by a set of traffic lights and a button for the pedestrian to press in order to request to cross. Once vehicles have stopped at the red light, the pedestrian will observe a green man to indicate it is safe to cross. After a period the green man will begin to flash, at which time vehicles will see a flashing amber light. As long as the crossing is now free of pedestrians vehicles may move on. (ii) The *Puffin Crossing* acts similar to a pelican crossing with the addition of sensors on the traffic lights and pavement and will return the traffic lights from red back to green once the pedestrian has completed their crossing. (iii) The *Toucan Crossing* is designed so that cyclists can also cross without dismounting, which is something they are required to do at any other type of crossing. These crossings are differentiated by the addition of a green bicycle alongside the green man. (iv) The *Equestrian Crossing* elevates the pedestrian button to make it accessible without the need to dismount in areas frequented by horse riders. For *uncontrolled crossings* the central island operates to split one crossing into two separate smaller crossings. However, unless a *controlled crossing* is staggered with separate pedestrian buttons on the central island, it should continue to be treated as one contiguous crossing. Road Rules 191-199 arise from requirements prescribed by *The Zebra, Pelican and Puffin Pedestrian Crossings Regulations and General Directions* 1997 (ZPPPCRGD) regulations 18 and 20, the *Road Traffic Regulation Act* 1984 (RTRA) section 25(5), and *The Traffic Signs Regulations and General Directions* 2002 (TSRGD) regulations 10, 27 and 28. Table 13 provides the structured English logic for Road Rules 191-199 and Table 14 provides the Boolean logic and equations, while Figure 15 provides a simple lawmap describing their operation.

**Technology Response:** The Mitsubishi[49], Vauxhall[50] and BMW[51] vehicles all included a pedestrian detection system that used the forward-looking windscreen-mounted camera and radar to sense pedestrians in the path of the vehicle. Manufacturers claim their systems will use audible and visual prompts to alert drivers to a pedestrian in the path of the vehicle, and will automatically apply emergency braking if the driver fails to respond by either altering the vehicle's direction of travel or braking to avoid an imminent collision.

---

[48] https://www.gov.uk/guidance/the-highway-code/using-the-road-159-to-203
[49] https://www.mitsubishi-motors.co.uk/cars/shogun-sport-shogun-sport/safety
[50] https://gb-media.vauxhall.co.uk/en-gb/07-05-insignia
[51] https://preview.thenewsmarket.com/Previews/NCAP/DocumentAssets/358227_v2.pdf

**Experimental Findings:** While the BMW in some instances identified the Give Way (red bordered white triangle) sign if present prior to an uncontrolled crossing, no vehicle actually identified the presence of a pedestrian crossing. The Mitsubishi and Vauxhall vehicles in most instances correctly identified pedestrians in both simulated[52] and real-world conditions[53]. However, in every instance the BMW vehicle either did not identify that a pedestrian was in its path, or in 30% of instances it incorrectly identified that pedestrian as a stationary small vehicle on the display in the instrument cluster. The Mitsubishi vehicle was the earliest of the three to brake, with test driver and passengers all suggesting it tended towards overreaction by braking too early and too far away from the pedestrian, and often in circumstances where even if the vehicle's speed had not been altered the pedestrian would no longer have been in the vehicle's path when the vehicle got there. It was also suggested that the Mitsubishi's approach to emergency braking in these situations was too severe, causing discomfort for vehicle occupants even when there was no perceivable risk to the pedestrian or vehicle. The Vauxhall alerted the driver audibly and with an image of a red person in the heads up display, but rarely engaged the brakes until it was too late to avoid some contact with the simulated pedestrian. Likely due to the fact that we never saw the BMW correctly identify a pedestrian even once, the BMW either failed to brake at all in situations where it did not see the pedestrian, or alerted using a red car shape in the heads up display and applied the emergency brakes in those situations where it incorrectly identified the pedestrian as a small vehicle. The failure of BMW's *Pedestrian Warning with City Brake Application* feature is not new, with our findings being similar to results released by the United States Institute for Highway Safety in 2019[54].

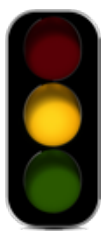

*Table 13: Structured English logic for Rules 191-199*

```
IF:
  [A] Where a vehicle is approaching a pedestrian crossing;

EXCEPT:
  [C] Where there is no person on or entering the pedestrian crossing, and:
             a. a flashing amber light;
             b. a green light; or,
             c. no traffic lights.

THEN:
  [X] The vehicle may proceed through the pedestrian crossing with caution.

ELSE:
  [Y] The vehicle must stop at the pedestrian crossing until there is:
             a. no person on or entering the pedestrian crossing; and,
             b. if present, the traffic light has changed to:
                    i. flashing amber; or,
                    ii. green.
```

[52] We used a similar approach to the United States Institute for Highway Safety in 37, using a padded mannequin as a simulated pedestrian both standing stationary and walking across the path of the vehicle travelling at 20mph (30kph).

[53] In real-world situations we stopped the vehicle at a crossing as required by the Road Rules. As we approached a pedestrian crossing, we watched for the vehicle to display a pedestrian icon in the instrument cluster or heads up display as described in the vehicle manufacturer's description of their pedestrian warning system. We also asked drivers to report serendipitous instances where a pedestrian stepped into the road in front of the moving vehicle. At no time were real pedestrians put in any risk of harm.

[54] https://metro.co.uk/2019/03/07/bmws-pedestrian-detection-system-works-really-really-badly-8835917/

*Table 14: Boolean Logic and Boolean Equations for Road Rule 103*

| Boolean Logic | Boolean Equations |
|---|---|
| IF A and C<br>THEN X<br>ELSE IF A and NOT C<br>THEN Y | X = A × C<br>Y = A × ~C |

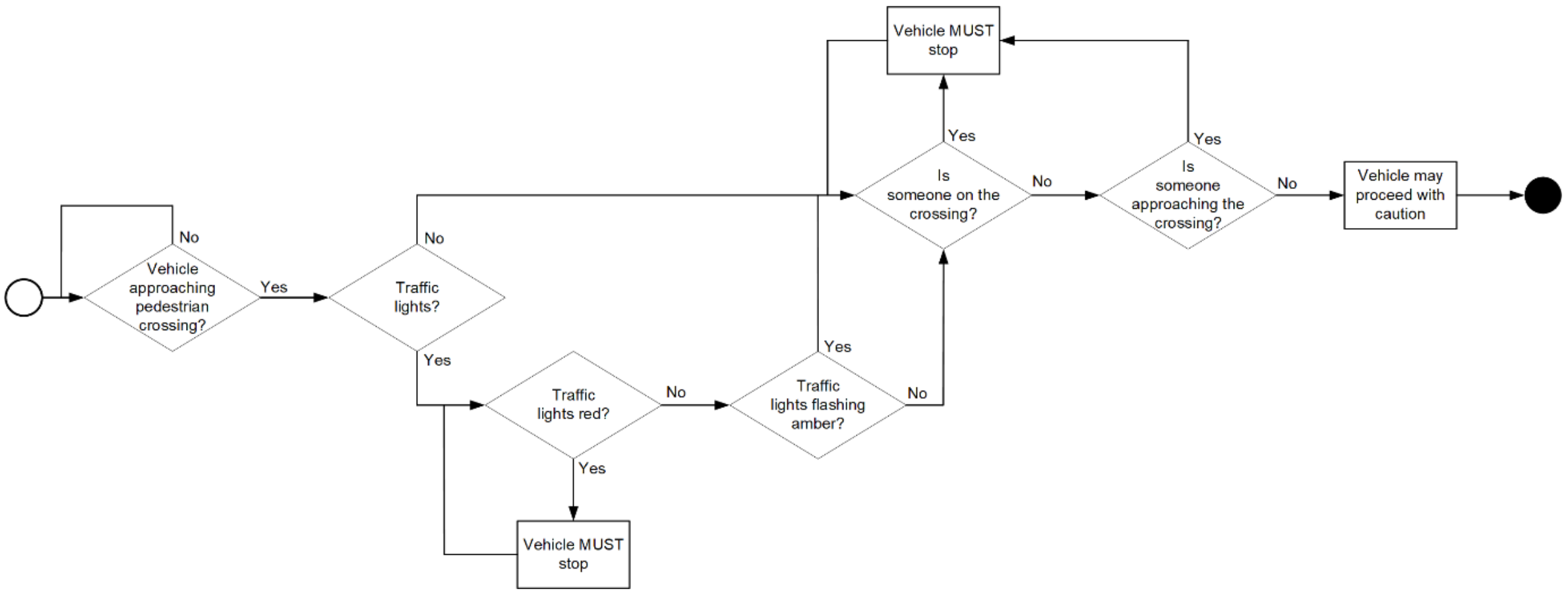

*Figure 15: Lawmap for Road Rules 191-199*

## (f) Rule 229

**Rule Description:** Rule 229[55] requires that before you commence a journey you **must** clear snow and ice from and demist all windows and mirrors and ensure that lights and number plates are clearly visible and legible. This requirement aggregates from a number of legislative sources, including: *The Road Vehicles (Construction and Use) Regulations* 1996 (RVCUR) s30(3)[56]; *The Road Vehicles Lighting Regulations* 1989 (RVLR) s23[57]; *Vehicle Excise and Registration Act* 1994 (VERA) s43[58]; and *The Road Vehicles (Display of Registration Marks) Regulations* 2001 (RVDRMR) s11[59]. Rule 229 also requires clearing of any snow and ice that might fall into the path of other road users: for example, snow that has collected on the roof of your vehicle that at speed might slide off and into the path of following traffic, or on braking could slide forward and down your windscreen obscuring your view[60]. Finally, the

---

[55] https://www.gov.uk/guidance/the-highway-code/driving-in-adverse-weather-conditions-226-to-237

[56] Section 30(3) states that all glass or other transparent material fitted to a motor vehicle shall be maintained in such condition that it does not obscure the vision of the driver while the vehicle is being driven on the road.

[57] Section 23(1) provides a requirement for all devices described in the paragraph to be both in *good working order* and *clean*, and s23(2) describes applicable devices to include headlamps, registration plate lamps, front position and side marker lamps, fog lamps, reflectors, stop lamps, running lamps, dim-dip devices, hazard lamps and headlamp levelling devices.

[58] Section 43 makes it an offence for any required registration mark such as the number plate fixed to a vehicle to be obscured or not easily distinguishable.

[59] Section 11 provides that it is an offence to affix reflex-reflective material to the number plate or to cause the letters to become retroreflective, or to use fixing devices such as screws in such a way that they change the appearance or legibility of the registration plate.

[60] Snow on the roof of a vehicle while driving could result in a £60 fine and could see the person in charge of the vehicle lose 3 demerit points from their driving license if observed falling off by an officer. Where that snow is

rule compels a check that the planned route is clear of any delays and that no further snowfalls or severe weather are predicted.

**Technology Response:** Many vehicle manufacturers now provide over-the-air (OTA) access to real-time traffic information that can alert drivers to potential hazards and congestion on their planned route, and even suggest alternate routes to avoid those issues. The Vauxhall and BMW vehicles both provided traffic alerts and updates as part of the navigation system[61], while the Mitsubishi vehicle only provided the public traffic alerts as part of the digital audio broadcast (DAB) system. However, no existing ADAS-enabled or semi-autonomous vehicles can self-identify or autonomously resolve many of the other requirements of this rule. It could be argued that requirements to de-ice and demist the windows might be unnecessary for SDV: that it is sufficient only for the SDV system to be capable of verifying that the cameras and any ultrasonic, radar or lidar sensors supporting the automated or autonomous functions are free of obstruction, interference or error, and where an issue is identified, alert and require the occupant to resolve the problem prior to commencing the journey.

**Experimental Findings:** The Vauxhall traffic alert tool was limited to providing delayed notice of traffic accidents and major incidents. Only the BMW tool was able to extend on this to provide near real-time traffic congestion detail by colour-coding roads both on the selected route, and in areas around the vehicle's current location and path. In testing, several installed ADAS technologies were capable of identifying and refusing to engage where frost, snow and other deliberately applied obstructions were present over the system's camera, ultrasound or lidar devices. Many were also described by the manufacturer as being capable of identifying errors and malfunctions, and when we introduced faults by disconnecting sensor or camera cables the vehicles correctly identified these issues with appropriate fault codes, and in most cases fully disabled the affected ADAS. However, no single vehicle system was able to identify and alert occupants to instances where the roof, headlights or registration plates were obstructed such that even the advanced vehicles we were using would allow driving in breach of these requirements of Rule 229.

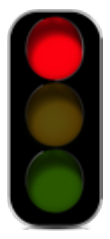

## 4.2 Summary of findings

Table 15 summarises results of the practical evaluation. These results show that even the most recent and advanced vehicle, the SAE Level 3 BMW 740Li, was unable to meet half of the requirements identified from the road rules in our test. While, as our traffic light rating indicates, many of the identified shortcomings may eventually be rectified with additional development of the software underpinning the observed to fall off onto your windscreen or onto another vehicle, an on the spot charge of *driving without reasonable consideration for other road users* which carries a minimum £100 fine and up to 9 demerit points may arise. In the most severe cases (such as https://www.bbc.co.uk/news/uk-scotland-tayside-central-55950942), or where either go to or are contested in court, the charge can escalate to *dangerous driving* and the fine may rise to as much as £1000.

[61] While Vauxhall provided a traffic alert system as part of ownership of the vehicle, the BMW traffic alert system is provided for only 2 years as part of the original vehicle purchase, and as a user-pays subscription service beyond that initial period.

smart automotive system or the inclusion of new or better trained artificial intelligence, it is possible some elements may remain unresolvable.

*Table 15: Practical evaluation results*

| Road Rule/s | | | Vauxhall Insignia | Mitsubishi Shogun Sport | BMW 740Li |
|---|---|---|---|---|---|
| **99-100** | 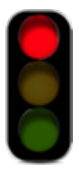 | Identifies when a restraint is required | ✓ | ✓ | ✓ |
| | | Identifies correct restraint type | ✗ | ✗ | ✗ |
| | | Identifies when any restraint is unplugged | ✓ | ✗ | ✓ |
| | | Cease smart function when unplugged | ✗ | ✗ | ✗ |
| **103-105** | 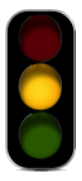 | Smart function reads most speed limit signs | ✓ | N/A | ✓ |
| | | Smart function identifies most give way signs | ✗ | N/A | ✓ |
| | | Smart function identifies most stop signs | ✗ | N/A | ✗ |
| | | Smart function is adherent to sign's instructions | ✗ | N/A | ✗ |
| | | Smart function automatically alerts when changing lane (LCA) | N/A | N/A | ✓ |
| | | Smart function automatically alerts when braking (ACC) | N/A | ✓ | ✓ |
| | | Smart function automatically cancels signal after use | N/A | ✓ | ✓ |
| | | Smart function can detect and cancel signal if it may be misleading to other road users | ✗ | ✗ | ✗ |
| | | Smart function recognises signals of other vehicles | ✗ | ✗ | ✗ |
| **113** | 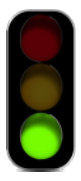 | Smart function identifies and responds to low light conditions by activating tail, plate and head lights | ✓ | ✓ | ✓ |
| **127-132** | 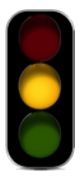 | Smart function identifies lane markings | ✓ | N/A | ✓ |
| | | Smart function alerts driver when about to cross lane markings | ✓ | N/A | ✓ |
| | | Smart function able to keep vehicle 'in lane' | ✓ | N/A | ✓ |
| | | Smart function prevents crossing solid double lines | ✗ | N/A | ✗ |
| **137-138** | 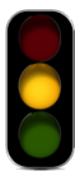 | Smart function correctly identifies when vehicle is not, but should be, in left-most lane | ✗ | N/A | ✗ |
| **191-199** | 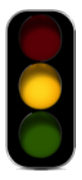 | Smart function correctly identifies most pedestrians | ✓ | ✓ | ✗ |
| | | Smart function takes appropriate action to avoid accident with pedestrian | ✗ | ✓ | ✗ |
| | | Smart function identifies pedestrian crossings | ✗ | ✗ | ✗ |
| **229** | 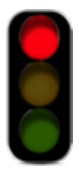 | Smart function able to verify snow and ice cleared from vehicle | ✗ | ✗ | ✗ |
| | | Smart function able to verify windscreen is free of snow and ice and demisted | ✓ | N/A | ✓ |
| | | Smart function able to verify lights and number plates are visible and free of obstruction | ✗ | ✗ | ✗ |
| | | Smart function provides traffic incident and accident alerts | ✓ | N/A | ✓ |
| | | Smart function provides information on traffic congestion on route | ✗ | N/A | ✓ |
| | | Smart function suggests routes to avoid incidents and/or congestion | ✓ | N/A | ✓ |

✓ The ADAS or autonomous smart function consistently met this requirement
✗ The ADAS or autonomous smart function consistently failed to meet this requirement
N/A The vehicle was not fitted with a relevant ADAS or autonomous smart function

# 5. Discussion

Most autonomous cars literature in the legal domain discuss either: (i) the issue of legal liability; (ii) novel privacy issues due to the vast amounts of data collected by vehicle systems and the resulting impact autonomous vehicles may have on owners and occupants freedoms; or both[62]. A strong argument has also been made that the two issues are interrelated and that in order to resolve one, legislation must also be cognisant of the other[63].

Most modern vehicles, but especially those with advanced ADAS or autonomous functionality operate as a collection of interconnected computing systems generating, sharing and consuming vast amounts of data. This data can include current and frequented locations, personal information about the owner or driver[64], and details about the vehicle's configuration and present serviceable status[65]. However, the data retained by many vehicles in so-called black box *event data recorders* (EDR) can also include the configuration and disposition of many of the vehicle's systems, some history of driving habits that can include speeding, excessive use of brakes and high g-force manoeuvres, whether safety restraints were in use, recent collisions and near misses, and in some cases video of these events[66]. Issues regarding manufacturer, insurer and police access to the data retained within an individual's vehicle have already come before the courts[67], and while much has been written about the fact that many vehicle owners and users remain unaware of the EDR and the data it collects, in all cases to date until it was downloaded the data was at least still physically located within their vehicle.

When the law says you should not drive without wearing your seat belt or that you should not exceed the posted speed limit, questions arise regarding the data collected and stored by the vehicle: whether the driver and occupants can be said to have provided informed consent for data collection[68], whether they have access to review and request correction or removal, and whether such collection and storage was legal in the context of existing data protection laws. Vehicles like the Vauxhall and BMW models reviewed in this work only increase the potential for privacy and data protection issues as a direct result of their *always-on* connected nature. Ever-increasing subsets of that data are now transmitted to manufacturer servers for re-transmission to manufacturer-branded smartphone apps and in order to provide training and usage data to improve ADAS and autonomous functionality. Not only does the potential for privacy breaches increase once the data leaves the vehicle, but in some jurisdictions questions may arise regarding at what point the subject who caused the data to be created ceases to retain ownership over data that could: (a) potentially expose them to prosecution for minor (or potentially major) traffic offences[69] that might normally have remained unknown; or (b) put them or their family at risk[70].

---

[62] Boeglin, J. (2015). The costs of self-driving cars: reconciling freedom and privacy with tort liability in autonomous vehicle regulation. *Yale JL & Tech.*, *17*, 171 at 175.
[63] *Ibid*.
[64] Some vehicles now provide a driver profiling platform, allowing the driver and frequent users to create a 'driver profile' that can be accessed using a pin number entered on the infotainment screen, a smartphone app or be linked to the key fob carried by the driver. Profiles can contain the person's name, vehicle settings and preferences, and smartphone details. In some internet-connected vehicles the profile can email log book style details of trips to the driver's email account.
[65] The BMW vehicle can send details about the current configuration of the ventilation system, current location, 360 degree camera views around the vehicle and even the remaining fuel volume and mileage to BMW's web servers, from where it can be shared to the BMW apps on any smartphone device that has been paired with the vehicle's VIN number.
[66] The Vauxhall vehicle's documentation included description of a data recorder that retained "the last 30 minutes of driving". The BMW vehicle is capable of recording and transmitting a wide variety of data regarding the driving conditions, driver inputs and even dashcam style video from the BMW Driver Recorder iDrive infotainment app. Both vehicles were configured and capable of using their respective cellular connectivity to alert a vehicle management call centre with severity (whether and how many airbags had triggered) and location data in the event of an accident.
[67] For example: *State of Florida v Worsham* [2017] FDCA 4D15-2733 where Worsham had had an accident that killed his passenger. Police had downloaded data from the event data recorder before seeking the search warrant; which was denied on the grounds that the search had already occurred. The Florida District Court of Appeal ruled the data inadmissible as the police had taken it without Worsham's approval or a valid search warrant. Also: Antonio Boparan Singh first admitted and was later convicted of causing serious injury by dangerous driving when he crashed into another vehicle causing permanent disabling injuries to a 1-year old child who later died. Singh only admitted culpability after black box data from his expensive high performance V8 SUV showed he had been driving at almost 2.5 times the legal speed limit on a residential street at the time of the accident.

Some promote the idea of future product infallibility, claiming that significantly improved reliability and safety should justify shifting all liability away from the manufacturer onto the autonomous vehicle owner through an application of strict liability[71]. However, such a solution creates difficulties for owners, insurers and courts and would only increase overall costs and amplify existing litigation tensions where the owner must square off against the insurer, and jointly or severally they are then required to go up against the manufacturer in cases where an alleged autonomous systems malfunction may have been at fault. It could be many years before victims or the driver are *made whole*. A more reasoned approach states liability should rest with the vehicle, or the human occupant sitting in what is ostensibly still referred to as *the driver's seat*, depending on their degree of physical capacity and mental alertness at the time of the accident[72]. In an assessment based on consideration of a similar fact accident scenario[73], Gurney suggests that if the human occupant is disabled that manufacturer liability should be automatic. If the human occupant can be described as having some form of diminished capacity or was distracted and fully reliant on the autonomous system at the time of the accident, then the manufacturer would bear at least partial liability. However, he argues that even when the autonomous system is in full control, the human occupant who appears attentive should retain sole liability.

The question of responsibility and resulting liability truly arises only when the autonomous system, rather than human occupant, is in full control of the driving task and something goes amiss. This is because most current law assumes strict liability based on the actions of a human driver[74]. If responsibility for the driving task shifts to the vehicle, then it would follow that the liability and the requirement to adhere to the road rules would also shift to the vehicle. However, such an assertion seems at odds with the fact that as we have seen in our test, many current road rules impose requirements that existing vehicles, and possibly even those in the foreseeable future, may be unable to fully address. Consider the example that arises out of Figure 11. If the autonomous system were engaged and was to act upon the erroneous input that led the TSR system to believe the posted speed limit was 80mph, this would see the vehicle driving at 50mph above the true speed limit in a restricted residential street. The potential for accident or significant injury to vehicle occupants, other road users and pedestrians cannot be overstated, and while it is the automated system that is at fault under our current laws the vehicle occupant in the drivers' seat would be held responsible. There is also the issue of other requirements that ae triggered only when a breach of some road rule has occurred[75], and whether autonomous vehicles will ever be truly capable of addressing these as well.

---

[68] Buhram, J. (2007). Riding with Little Brother: Striking a Better Balance between the Benefits of Automobile Event Data Recorders and Their Drawbacks. *Cornell JL & Pub. Pol'y*, *17*, 201 at 217; discusses the current lack of informed consent for EDR data collection and the risk that authorities might use that data as evidence against a driver in civil or criminal proceedings without the driver first having agreed to incur this legal risk.
[69] *Ibid*.
[70] Bevin, E. (2019). Man pleads guilty to stalking and controlling ex-girlfriend's car with his computer. *ABC News* (6 Nov 2019). Sourced from: https://www.abc.net.au/news/2019-11-06/ract-employee-pleads-guilty-to-using-app-to-stalk-ex-girlfriend/11678980. A man plead guilty and received a sentence of imprisonment after using details of the victim's car to facilitate access to data created by the victim that was stored in the car including location information regarding where she was and when, and in order to take remote control of her vehicle.
[71] 41 at 473-474.
[72] Gurney, J. K. (2013). Sue my car not me: Products liability and accidents involving autonomous vehicles. *U. Ill. JL Tech. & Pol'y*, 247.
[73] The person uses a Google fully autonomous vehicle to undertake a journey and for all but the attentive driver, the autonomous system suffers a malfunction.

It has been said that the issues arising from autonomous vehicles can be dealt with by existing legal frameworks and regulations[76]. The *monitoring and control tests* requirement for *compliance with relevant road traffic rules* in the UK *Department of Transport's* (DoT) response to the *Safe Use of Automated Lane Keeping System* (ALKS) *on GB roads*[77] reinforces not only the suggestion that existing road rules should be considered adequate, but also that autonomous systems must be adherent to them. However, wholesale import of our existing road rules as the model to govern autonomous systems may at best be unreasonable, and at worst could prove impossible. It may also create conflicts with other laws that would not arise where a human is in control of the driving task, including laws that may provide exemption from compliance with particular road rules in special circumstances[78]. Many have concluded that new regulation is likely necessary not only to address barriers to adoption[79], but also to address the issues of co-existence[80], liability[81] and whether and how individual road rules will apply to vehicles when the vehicle itself is in autonomous control of the driving task. However, when even the best of human drivers may not adhere to all road rules at all times, it may be especially challenging for autonomous vehicles to co-exist if their software rigidly enforces the requirements of every road rule without exception[82]. Co-existence requires the autonomous system be capable of a degree of flexibility in its decision-making when it encounters the unpredictable moving feast that is human behaviour and response. Our practical experiment demonstrates that existing SAE level 2 ADAS and level 3 autonomous system adherence to the complex requirements of some road rules can at best be variable, and our results would support the contention that holding these systems to strict observance of all current UK road rules will prove impossible. For exactly these reasons it is suggested that it may be necessary to develop and train autonomous systems with a capability for *strategic rule breaking*[83].

---

[74] Duffy, S. H., & Hopkins, J. P. (2013). Sit, stay, drive: The future of autonomous car liability. *SMU Sci. & Tech. L. Rev.*, *16*, 453.
[75] For example: the duty to stop and render assistance after an accident.
[76] Brodsky, J. S. (2016). Autonomous vehicle regulation: How an uncertain legal landscape may hit the brakes on self-driving cars. *Berkeley Technology Law Journal*, *31*(2), 851-878.
[77] https://assets.publishing.service.gov.uk/government/uploads/system/uploads/attachment_data/file/980644/Safe-Use-of-Automated-Lane-Keeping-System-ALKS-Call-for-Evidence-FINAL-accessible.pdf, para 5.1.2, p39.
[78] In many jurisdictions other legislation creates medical and disability exemptions for seatbelts. Any system that requires strict compliance with existing road rules must surely provide a human override in cases where such exemption exists.
[79] 34 at 853.
[80] Co-existence with existing non- and semi-autonomous vehicles.
[81] Much has been written to expose and discuss the myriad issues surrounding allocation of fault, apportionment of blame and liability when vehicles operating in autonomous control mode are involved in traffic accidents. Others have extended these arguments into the philosophical area of imbuing technology with ethical decision-making.
[82] Schimelman, B. I. (2016). How to Train a Criminal: Making a Fully Autonomous Vehicles Safe for Humans. *Conn. L. Rev.*, *49*, 327.

Another issue exists for human drivers that could be characterised either as overconfidence[84] or complacency[85]. Drivers will use the autonomous systems and become increasingly self-assured of the system's ability to safely perform the driving task. They will progressively develop risk tolerance and become increasingly more willing to allow themselves to be distracted while the autonomous system is in control of the vehicle. Proponents of the *risk homeostasis* theory[86] would suggest that as autonomous vehicle systems like Tesla's *Autopilot* and BMW's *Driver Assistant Professional* become more capable and potential increases in vehicle safety are realised, risk tolerant drivers will use the journey as some autonomous vehicle advocates intend: to catch up on work, read or watch movies. While distracted, these autonomous vehicle owners will not observe cues of an impending accident and will fail to take necessary corrective action, resulting in accidents and a rebound to pre-autonomous vehicle accident levels. Examples of this effect can already be seen where misplaced confidence in these newly developed autonomous systems is resulting in accidents arising from reckless and risk tolerant behaviours[87].

It has recently been suggested that the race to full SAE Level 5 automation is itself the issue: the reason existing philosophies and approaches for creating the safe fully autonomous vehicle have failed to truly realise that which they seek[88]. While the push for automation continues unabated and in many different directions, safety is being considered by technologists only after the primary function of the smart system has already been successfully resolved: often using data collected from large groups of early adoption users to understand and resolve safety-related issues[89]. The approach of many philosophers, ethicists and autonomous car researchers has been to further cloud the safety issue with fundamentally pointless no-win moral dilemmas based on the 'trolley problem'[90]. Barry Lunn and others argue that the safety issue should come first: that AI should be developed to learn from human drivers rather than be developed as empty vessels to replace them, and that improving systems and safety to avoid accidents so that we never have a trolley problem is preferable to continued philosophising about how we might wish an already established system to respond[91].

[83] *Ibid*.
[84] Innes-Jones, G., & Scandpower, L. (2012). Complacency as a Causal Factor in Accidents-Fact or Fallacy. *London: IChemE Symposium Series 158*, Hazards XXIII.
[85] *Ibid*. Complacency in this context is defined as: self-satisfaction which may result in non-vigilance based on an unjustified assumption of satisfactory system state.
[86] Wilde, G. J. (1989). Accident countermeasures and behavioural compensation: The position of risk homeostasis theory. *Journal of Occupational Accidents*, *10*(4), 267-292.
Risk homeostasis theory suggests that as new systems or features increase safety, people become more willing to engage in less safe behaviours, thus serving to maintain the accident or fatality levels at a static level. Examples include that as seat belts, airbags and anti-lock braking systems were implemented and cars became safer with fewer accidents, drivers became more willing to speed and drive too close to the vehicle in front, serving to increase accident and fatality levels and return them to previous levels.
[87] Examples include: (i) Devainder Goli of Raleigh in the USA who was charged after being observed watching a movie while Tesla Autopilot drove his car at highway speeds; (ii) Param Sharma of California who has been charged multiple times and had his vehicle confiscated after being seen mid-journey to climb from the drivers' seat into the back of his Tesla while Autopilot was engaged and the vehicle was in motion; (iii) Joshua Brown of Florida whose confidence in the autonomous system led him to watch a Harry Potter movie while Autopilot drove his Tesla into the trailer of a large semi truck, killing him; and (iv) at least three examples where Tesla vehicles in Autopilot mode drove into the backs of stationary fire trucks while occupants were engaged in reading, watching movies on their phones or other non-driving tasks.
[88] Dalvin Brown, "How should autonomous cars make life-or-death decisions? In the best of worlds, they won't". Washington Post (06th August, 2021). https://www.washingtonpost.com/technology/2021/08/06/self-driving-ai-death-decisions/

# 5. Conclusion

Our work shows it is possible to *deconstruct* traffic law into a series of structured rules that can then be used to *construct* tools: Boolean equations and visual representations. These tools can be used in development of AI and autonomous systems, and along with the structured rules, in verification and validation of the operation of the rule of law in the resulting system's decision-making processes and outcomes. We have observed on other projects[92] and in other domains[93] that this (de)construction approach and the resulting tools are within the capabilities of expert practitioners, and that the tools make often highly technical domain-specific knowledge sufficiently comprehensible for AI developers and decision scientists who seek to make legislation and the rule of law an active component of their AI decision-making models.

**Acknowledgements**

The authors acknowledge support for this research from the following sources: From Innovate UK under project *Engine-B: Powering an age of professional services* - 106158; EPSRC under project *AISEC: AI Secure and explainable by construction* - EP/T026952/1; and RAEng under project *SafeAIR: Safer aviation from ethical Autonomous Intelligence Regulation*.

[89] Stephen Edelstein, "Tesla's autonomous-car efforts use big data no other car maker has". Green Car reports (30th December, 2016). https://www.greencarreports.com/news/1108065_teslas-autonomous-car-efforts-use-big-data-no-other-carmaker-has

[90] Awad, E., Dsouza, S., Kim, R., Schulz, J., Henrich, J., Shariff, A., ... & Rahwan, I. (2018). The moral machine experiment. *Nature*, *563*(7729), 59-64.

[91] See 71.

[92] See 10.

[93] McLachlan, S., Paterson, H., Dube, K., Kyrimi, E., Dementiev, E., Neil, M., Daley, B., Hitman, G.A., & Fenton, N. (2020). *Real-time Online Probabilistic Medical Computation using Bayesian Networks* (No. 2744). *Proceedings of the IEEE International Conference on Health Informatics (ICHI 2020).* DOI: 10.1109/ICHI48887.2020.9374378; and, McLachlan, S., Kyrimi, E., Dube, K., & Fenton, N. (2019). Clinical caremap development: How can caremaps standardise care when they are not standardised? *Proceedings of the 12th International Joint Conference on Biomedical Systems and Technologies (BIOSTEC 2019)*, volume 5: HEALTHINF, pp 123-134. DOI: 10.5220/0007522601230134